\documentclass[electronic]{vgtc}             % electronic version

\ifpdf%                                % if we use pdflatex
  \pdfoutput=1\relax                   % create PDFs from pdfLaTeX
  \usepackage{graphicx}                % allow us to embed graphics files
  \DeclareGraphicsExtensions{.pdf,.png,.jpg,.jpeg} % for pdflatex we expect .pdf, .png, or .jpg files
\else%                                 % else we use pure latex
  \usepackage{graphicx}                % allow us to embed graphics files
  \DeclareGraphicsExtensions{.eps}     % for pure latex we expect eps files
\fi%

\graphicspath{{figures/}{pictures/}{images/}{./}} % where to search for the images

\usepackage{microtype}                 % use micro-typography (slightly more compact, better to read)
\PassOptionsToPackage{warn}{textcomp}  % to address font issues with \textrightarrow
\usepackage{textcomp}                  % use better special symbols
\usepackage{mathptmx}                  % use matching math font
\usepackage{times}                     % we use Times as the main font
\usepackage{cite}                      % needed to automatically sort the references
\usepackage{tabu}                      % only used for the table example
\usepackage{booktabs}                  % only used for the table example
\usepackage{color}
\usepackage{epstopdf}
\usepackage{amsmath,bm} % added by JY
\onlineid{1178}

\vgtccategory{Research}

\vgtcinsertpkg

\title{An Information-theoretic Visual Analysis Framework for Convolutional Neural Networks}

\author{Jingyi Shen\thanks{e-mail: shen.1250@osu.edu}\\ %
        \scriptsize The Ohio State University %
\and Han-Wei Shen\thanks{e-mail: shen.94@osu.edu}\\ %
     \scriptsize The Ohio State University %
}

\abstract{
Despite the great success of Convolutional Neural Networks (CNNs) in Computer Vision and Natural Language Processing, the working mechanism behind CNNs is still under extensive discussions and research. Driven by a strong demand for the theoretical explanation of neural networks, some researchers utilize information theory to provide insight into the black box model. However, to the best of our knowledge, employing information theory to quantitatively analyze and qualitatively visualize neural networks has not been extensively studied in the visualization community. In this paper, we combine information entropies and visualization techniques to shed light on how CNN works. Specifically, we first introduce a data model to organize the data that can be extracted from CNN models. Then we propose two ways to calculate entropy under different circumstances. To provide a fundamental understanding of the basic building blocks of CNNs (e.g., convolutional layers, pooling layers, normalization layers) from an information-theoretic perspective, we develop a visual analysis system, CNNSlicer. CNNSlicer allows users to interactively explore the amount of information changes inside the model. With case studies on the widely used benchmark datasets (MNIST and CIFAR-10), we demonstrate the effectiveness of our system in opening the blackbox of CNNs. 
} % end of abstract

\CCScatlist{
  \CCScatTwelve{Convolutional Neural Network}{Information Theory}{Entropy}{};
  \CCScatTwelve{}{}{}{}
}
\begin{document}


\firstsection{Introduction}
\maketitle
The Convolutional Neural Networks (CNNs) is a type of deep neural networks that have shown impressive breakthrough in many application domains such as computer vision, speech recognition and natural language processing \cite{Khan:2019:SurveyDCNN}. Different from the traditional multilayer perceptrons which only consist of fully connected layers, CNNs have additional building blocks such as the convolution layers, pooling layers, normalization layers, dropout layers, etc. Through a combination of these layers, a CNN model can extract features of different levels for generating the final inference result.  Despite its great success, one question often raised by the deep learning researchers and users is: ``how and what does the network learn?'' Because of the lack of theoretical explanation of CNNs, designing and evaluating the model is still a challenge task.

The interpretation of the black box systems in deep neural networks (DNN) models has received a lot of attention lately. Researchers proposed various visualization methods to interpret DNNs such as CNNVis \cite{Liu:2017:CNNVis}, LSTMVis \cite{Strobelt:2018:LSTMVis}, GANViz \cite{Wang:2018:GANViz}, etc. By making the learning and decision making process more transparent, the reliability of the model can be better confirmed which allows researchers to improve and diagnose the models more effectively. In 2017, the work by Shwartz-Ziv and Tishby \cite{Shwartz-Ziv:2017:OBBDNNVI} tackle the problem using information theory to analyze the learning process of DNNs via a technique called {\em Information Plane}. By analyzing the mutual information between layers during training, the author observed that DNNs aim to first compress the input into a compact representation and then learn to increase the model's generalizability by forgetting some information. 

Since a CNN has the ability to reduce the input set into a more compact representation before the inference is made, it can be thought of as an information distillation process. Measuring the change of information from the input during this information distillation process makes it possible to open the black box of neural network models. Although plenty of works have been done to visualize DNNs, to the best of our knowledge, using information theory as a comprehensive analytic tool to visualize and analyze deep learning models has not been fully studied in the visualization community. In this paper, we aim to bring together information theory and visual analytics for analyzing deep learning CNN models. 

We take an information theoretical approach to better understand the information distillation process of CNNs. Starting with an overview of the model, we treat the model as a black box and analyze its input and output. Then we get into the details of the model's building blocks such as layers and channels. There are multiple angles to assess the information, e.g., measure the amount of information inside the input data, between the input and output, between the intermediate layers and among the channels. To systematically formulate the queries, we introduce a data model in the form of a 4-dimensional hypercube that allows users to systematically query the data from  various stages of CNN models. The dimensions of this data model represent the input, layers, channels and training epochs. Different slicing or aggregation operations on this hypercube are introduced to glean insights into the model.  To calculate the information flow through CNN models, we propose two different types of entropy: inter-sample entropy and intra-sample entropy. Inter-sample entropy is the entropy from a set of samples, including the input and the intermediate results, in their original high-dimensional space, and intra-sample entropy is the entropy for each individual data sample such as an image or a feature map. We design and develop a visual analysis system, CNNSlicer. 
%providing evidence and interpretations of the performance, and conducting comparative analysis with the evidence.  
CNNSlicer aims to reveal the information inside the model and help users understand the learning process of a CNN model or evaluate a well-trained CNN model in the testing phase.  
 
 %The first component is the CNN architecture view, including all building blocks of this CNN. The data model view illustrates the query under interest. The training performance view, the layer view and the channel view offer interpretations of the model from different perspectives, such as how much information get filtered between layers and which channel contributes the most for the classification task. Additionally, we use several case studies to demonstrate the functionality of our system. 
 
% to expose the features that different convolutional filters extracted.\
%  associated decision-critical data in the structures (e.g., neurons, activations, parameters, etc.). 
% to visually analyze the structured information from DNNs
In summary, the contributions of our work are: 
(1) We propose a novel hypercube data model to help users make more systematic queries to the information available in CNN models.
(2) We propose two types of entropy, inter-sample entropy and intra-sample entropy, each reveal different insight into the information distillation process of CNNs. 
(3) We combine visual analysis of CNNs with information theory and develop a system, CNNSlicer that allows users to perform visual queries into the CNN models. 

\section{Related work}
\subsection{Information Theory}
Information theory, first introduced by Claude Shannon in 1948, is a theory focusing on information quantification and communication \cite{IT1948Shannon}. Shannon also defined several fundamental concepts such as entropy, relative entropy, redundancy and channel capacity to measure information in a statistical way \cite{IT1948Shannon}. Among them, entropy, which quantifies the information, choice, and uncertainty \cite{IT1948Shannon}, is the most fundamental measurement. Even to date, information theory is still popular in areas like mathematics, computer science (e.g., feature selection \cite{Huang:2020:ITUFSHD}), electrical engineering (e.g., compression \cite{Huang:2020:ITUFSHD}), neuroscience, analytical chemistry \cite{FloresGallegos2013ProvisionalCS}, molecular biology (e.g., molecular communication \cite{Gohari:2016:ITMC} and biological sequence analysis \cite{Susana:2013:IT4bio}), etc. In our work, we want to utilize information theory as an measurement of information and some visualization methods to 
open the black box of CNNs.

\subsection{Visual Analytics for Deep Learning}
In both visualization and machine learning fields, an increasing number of researchers are now focusing on developing methods to allow visual analysis of deep learning models to diagnose, evaluate, understand and even refine the model. In visualization community, scientists integrate multiple visual components with proper interactions to let user explore and understand the model. For example, CNNVis \cite{Liu:CNNVis:2016} was proposed to assist users exploring and understanding the role of each neuron in the model. By only analyzing the input and output, Manifold \cite{Zhang:2019:Manifold} was designed to help the development and diagnosis of the model. To understand and interpret the Deep Q-Network (DQN), DQNViz \cite{Wang:2019:DQNViz} encodes details of the training process in the system for users to perform comprehensive analysis. In the machine learning field, there are some popular works on Explainable Artificial Intelligence (XAI) \cite{Arrieta:2019:XAI}, such as saliency map \cite{Fong:2017:IEBBMP}, loss landscape visualization \cite{Li:2017:lossLandscape}, sensitivity analysis \cite{Cortez:2011:SensitivityAnalysis} and deconvolution \cite{Zeiler:2013:deconvCNNvis}.

\section{Background} \label{background}
\subsection{Convolutional Neural Network} \label{CNN}
Different from the traditional multilayer perceptrons (MLPs) which only consist of fully connected layers, Convolutional Neural Networks (CNNs) have more flexibility with various building blocks including fully connected layers, convolutional layers, pooling layers, non-linear activations, dropout, normalization, short-cut, residual blocks, etc. The typical architecture of a CNN model usually starts with a convolutional layer which can extract local spatial correlations. After that, a non-linear activation function is applied to learn the non-linearity of the data and generate what we call activations. The next operation is normally a pooling layer which helps to downsample the features and makes it less sensitive to spatial translations. With the above layers stacked up multiple times, CNNs are able to automatically extract hierarchical features from the input. 
% One powerful thing about CNNs is that they largely reduce the number of parameters to be optimized through weight sharing during convolution operations. 
For a classification task, normally there will be several fully connected layers at the end of the CNN architecture. These fully connected layers can combine and convert the extracted features into probabilities of classes \cite{Khan:2019:SurveyDCNN}. Besides these basic blocks, domain experts use dropout \cite{Srivastava:2014:Dropout} which can randomly drop network connections during training to address the overfitting problem. Batch normalization \cite{Ioffe:2015:BN} which acts like a regularizer, is utilized to stabilize and accelerate the training process.

% However, the above building blocks are often taken for granted in CNN architectures, with the theoretical explanation of them being ignored. 
Consider a standard CNN model as an information distillation process, each layer is basically extracting and purifying the input information into concise representations. Initially, the model has no idea what to filter out given the large number of input samples, where the information distillation is more like a random process. During training, the model's weights get updated through backpropagation and its output converges to the expected result. When a model is well trained, it can reduce and summarize the input by keeping only the most salient characteristics of the data to perform the underlying inference task. This fundamental nature of CNNs is what motivate us to develop a framework to  evaluate and explain how a neural network model process information and how the information flows through the model. % a neural network model through the amount of information
% Actually, there has been some attempt to open the black box via information theory. One of the pioneer work is by Tishby and Zaslavsky \cite{Tishby:2015:DLIBP}. In their work, they consider a basic layered neural network model as a Markov Chain with every layer only depends on the output of the previous layer. They show that the mutual information between hidden layers and input layers can provide quantification of the DNN. However, they did not put much effort on demonstrating their findings vividly, instead their goal is to give a theoretic bound of the DNN in an informatics perspective. The goal of our paper is to shed light on the black box model with the help of information theory and visualization. 
% Information-theoretic

\subsection{Information Theory} \label{information_theory}
% \section{Information Content Measurement of a Convolutional Neural Network Model} 
In information theory, entropy (Shannon's entropy \cite{IT1948Shannon} to be more specific) is a widely used metric to measure the amount of disorder (information) in the system. 

During a random process, if an event is more common, the occurrence of this event contains less information than a rare event. This is to say, given a random event, higher probability means lower information content. From this intuition, the information content of a stochastic event $x$ can be defined as the negative log of its probability $p(x)$:
\begin{equation} \label{eq:IX}
I(x) = -log_b(p(x))
\end{equation}
In Shannon's entropy \cite{IT1948Shannon}, the logarithmic base of the log is 2, so the resulting unit is "bit". 

For a stochastic system, if each random event can happen with almost the same probability, this system is almost unpredictable. From the information theoretic point of view, the system is more disorder. On the other hand, when only a few events are more likely to happen, the output will be less surprising and hence the system contains less information. Thus, the amount of information for a system, also known as the entropy \cite{IT1948Shannon}, is defined as the expected value of all random events' information content:
\begin{equation} \label{eq:Entropy}
H(X) = E[I(X)]= E[-log_b(P(X))] = -\sum_{i=1}^n P(x_i)log_bP(x_i)
\end{equation}
where $n$ is the number of possible stochastic events of the system. 

\begin{figure}[htp]
    \centering
    \includegraphics[width=9cm]{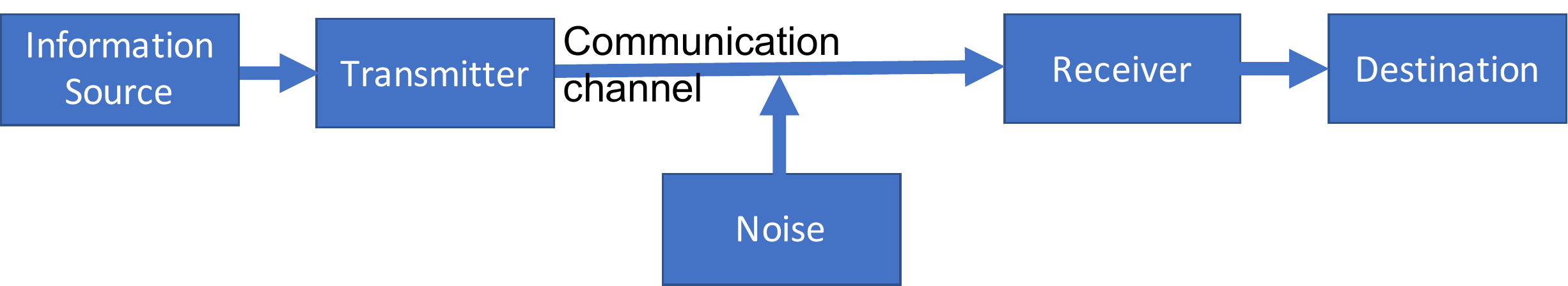}
    \vspace{-9pt}
    \caption{A communication system transmits encoded source information to the destination through a noisy communication channel.}
    \label{fig:com_system}
\end{figure}

In \cite{IT1948Shannon}, Shannon also defined a theoretical upper bound for a communication system. The system has six components as shown in \autoref{fig:com_system}. The {\em information source} produces the original information to be transmitted to the destination. A {\em transmitter} encodes the information into a suitable representation that can be transmitted from the transmitter to the {\em receiver} through the {\em communication channel}. The receiver is a decoder which decodes the received information and send it to the {\em destination}. During transmission, the addition of {\em channel noise} will cause signal interference. Shannon defined the capacity of a noisy channel by \cite{IT1948Shannon}:
  \begin{equation} \label{eq:CC}
  Capacity = H(X) - H(X|Y)
  \end{equation}
where $H(X)$ measures the amount of information in the information source and the conditional entropy $H(X|Y)$ gives us the amount of uncertainty for the source information $X$ given the destination information $Y$. In other words, $H(X|Y)$ measures the the amount of information loss during the transmission. If $Y$ has exactly the same amount of information as $X$, then given $Y$ there would be no uncertainty about $X$. In this case, $H(X|Y)$ equals to zero. However, as we have inevitable noise added during transmission, in practice $H(X|Y)$ is not zero.

\subsection{Information Theory in Deep Learning}
Information theory has been employed in deep learning. In information theory, cross entropy measures how different  two probability distributions are. In deep learning, the cross entropy loss, which aims to minimize the distance between the predicted labels and the true labels, is widely used in classification tasks. However, information theory can be more powerful. 

Actually, there has been some attempt to open the black box of deep learning models via information theory. One of the pioneer work was done by Tishby and Zaslavsky \cite{Tishby:2015:DLIBP}. In their work, they consider a basic layered neural network model as a Markov Chain with every layer only depends on the output of the previous layer. They show that the mutual information between the hidden and input layers can quantify the characteristics of the neural networks. They did not put much effort to further their studies for practical applications, however. Instead, their goal is to give a theoretic bound of the neural networks in an information  perspective. The goal of our paper is to shed light on the black box model with the help of information theory and visualization. 

As we discuss in Section \ref{CNN}, CNNs can be seen as extracting and purifying the input information through consecutive layers. In information theory, entropy can measure the amount of information of a system. Since in our case the system is a neural network model, it would be interesting to know how the entropy changes inside the layered neural network. Since the information is  ``compressed'' into the output-relevant information, one may wonder whether the entropy keeps on decreasing as we get into deeper layers. On the other hand, through iterative training, the optimized model is more sure about what to output given an input. This means the randomness of the system is decreasing when making inferences. Can the entropy of the model reflect this change? Furthermore, how does the entropy of the feature maps from a specific layer change during training? If we can answer the above questions, we are able to improve the transparency of the CNN model through information theory. There are plenty of queries we can make. 
% {\color{red}{ To better organize our query and make it straightforward, we first propose a data model to represent the neural network. }}

% A well-trained classification model can transform the input data which contains a lot of irrelevant information into classes which has much less information. This transformation indicates that the model is actually converting high entropy data into low entropy. 
% hierarchical representations at the layered network naturally correspond to the structural phase transitions along the information curve. 

\section{A CNN Information Analysis Framework} 

\subsection{Requirement Analysis} \label{requirement}
The purpose of our work is to analyze and visualize CNN models from an information theoretic point of view. After thoroughly discussing with a machine learning and information theory expert, we come up with the following visual analysis requirements:
% After {\color{red}{discussed with machine learning and information theory experts}}, 
\begin{itemize}
    \item \textbf{R1: Provide an overview as well as the basic building blocks of a CNN model.} CNN models have hierarchical structures. To go along with this property, our visual design also requires such a hierarchy. From the overview of the model to the details of the architecture, this top down visualization can assist users to understand the model in a more intuitive way. 
    
	\item \textbf{R2: Develop a unified data model for CNNs.} As we discuss before, there are various of information queries one can make to improve the transparency of the CNN model. Visualizing all possible queries in a theoretical way would help users to gain a deeper  understanding of the model. 
	However, there is a large amount of data generated during the training process and depends on the application, CNN architectures can vary. Thus, a unified CNN model representation is needed. 
% used for specifying the training progress and diagnosing training failure. However, visualizing all the training dynamics will cause severe visual clutter.

	\item \textbf{R3: Visualize information statistics of the model.} To give a clue about how the CNN model makes decisions from an information theoretic perspective, we need to analyze the statistics associated with each building block of the model. However, since there is a huge amount of statistics information to show, we need better data organization and effective visual designs to avoid visual clutter. With detailed statistics about the flow of information, we are able to find interesting directions for further analysis and potential explanation for the model. 
% 	As we combine statistics with model's blocks, we are able to find some {\color{red}{interesting entry points}}

% 	\item \textbf{R4: Support the analysis of CNN's training process.} Since the training is done through iteration. It is necessary that we take training epochs into consideration when trying to explain the learning process of the model. 
	
	\item \textbf{R4: Support multifaceted  analysis of the model's learned features.} Besides the statistical information, to get insight into the functionality of the intermediate layers, comprehensive visualization to display detailed multifaceted information in the model is required. For example, the visualization of the various features that the model learned. 
	
\end{itemize}

\subsection{dCNN: A Data Model for CNNs} \label{datamodel}
As we stated before, it is important to formulate the queries systematically to evaluate and interpret the information flow inside a CNN model. In this section, we formalize a data model to represent the information available in CNN models in a comprehensive way, regardless of the specific type of CNN architecture. This data model helps us organize the information queries and make the CNN explanation process more systematic. 

The data available in the entire CNN model can be thought of as a four dimensional array, denoted as $dCNN(X, L, C, T)$, where each dimension of the array serves as a key and a specific value of the key is used to slice the array and obtain the information stored within.

The first dimension of the array is the input data (training or testing) dimension, denoted as $X$. Within this dimension, each specific instance, or a value in the dimension represents an input, for example, an input training image. In the case of classification neural networks, the dimension $X$ can be further divided into subgroups where each subgroup represents data in a specific class.
 
The second dimension in the multidimensional array is the layer dimension, denoted as $L$. As we know, a convolutional neural network contains multiple layers, from the input to hidden to the output layers, and each of the layers plays a specific role. For example, the earlier layers are responsible for extracting low level features such as edges and colors. As we go deeper into the model, low level features are combined into high level features such as contours or textures and then objects can be extracted. As each layer acts like a `function' whose output only dependents on the input from the previous layer, the input information get distilled layer by layer. To understand how the information flows through the model, it requires us to query the information shared between layers.

Given a particular layer in the CNN, there exist multiple convolution kernels, also known as filters and the output of which is often called a channel. If we fix at a layer, there can be multiple channels to be chosen for analysis, which prompts the necessity of having the next dimension in the array that represents the channels, denoted as $C$. For any CNN model, it is crucial to have multiple filters to be trained, and as a result multiple feature maps will be produced for a given input. Intuitively, each channel is activated by some specific features in the input. It is also known that more filters does not necessarily guarantee more information from the input to be captured since some filters may be `dead' where no information is extracted. Since the capability of a filter can be checked by the corresponding feature maps, it is reasonable that we focus on feature maps when evaluating the filters.

Finally, a CNN model is trained through many iterations of backpropagation and gradient descent. When the input training data are divided into many small batches, called mini-batches, a backpropagation is conducted for each mini-batch and an iteration that goes through all mini-batches is called an epoch. To track the progress of training for a CNN model over time, we index the data in the last dimension of the four dimensional array by its epoch number, denoted as $T$.  
% \begin{figure}[htp]
%     \centering
%     \includegraphics[width=5cm]{pictures/4d.pdf}
%     \vspace{-9pt}
%     \caption{.}
%     \label{fig:4d}
% \end{figure}

As a result, we propose to use a 4-dimensional hypercube to represent the data related to a CNN model for model evaluation. The four dimensions are: data ($X$), layer ($L$), channel ($C$) and training epoch ($T$). We denote this data structure as $dCNN(X, L, C, T)$. We note Inside the four dimensional array, each specific entry is often an array, for example, an input image, a feature map in the hidden layer, or an array of probability values, one for each possible output label. Different slicing or aggregation operations on this hypercube lead to different information queries and facilitate the evaluation and interpretation of the CNN model.

% Note that the channel dimension is conceptually independent but actually dependent on which layer we are looking into. 

\subsection{Slicing dCNN as Information Query} {\label{entropy}}
The data model described in Section \ref{datamodel} defines a 4-dimensional hypercube to represent all data that can be extracted from a CNN model. To analyze the data in a CNN mode, we can first slice the hypercube based on the analysis need, followed by calculating various entropy measures from the slicing result. Below we first explain the meaning of data slicing. The calculation of entropies is explained in Section \ref{EntropyCal}. 
 \vspace{-6pt}
\begin{figure}[htp]
    \centering
    \includegraphics[width=5cm]{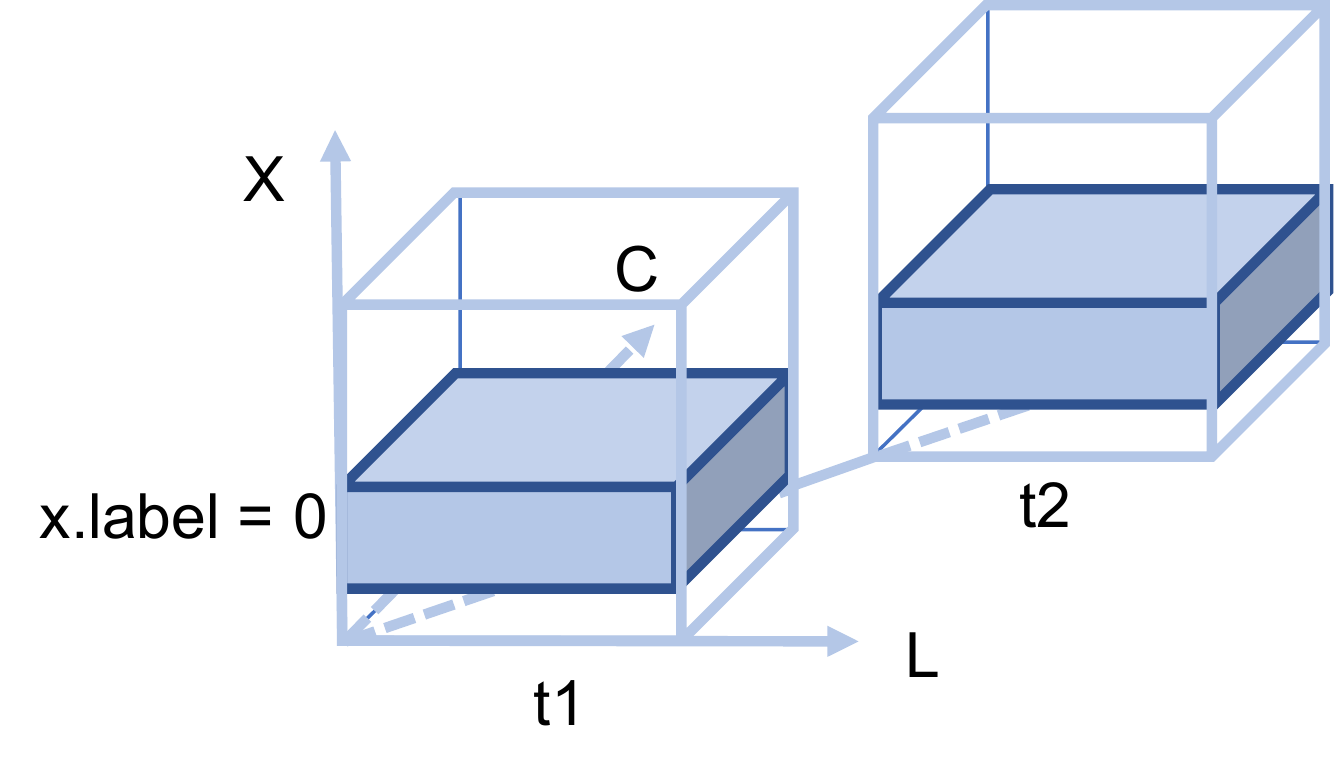}
    \vspace{-9pt}
    \caption{Slice the data model by input (X) dimension to get all intermediate data in the CNN model, given input with a class label 0.}
    \label{fig:xlabel0}
\end{figure}

\begin{itemize}
\item {\bf Slicing by Input $X$}: This is denoted as $dCNN(x, -, -, -)$, which means we slice the hypercube on the $X$ dimension where $x$ is a particular instance or a set of instances, and the notation $-$ indicates all values in the respective dimension. As an example, $dCNN(x, -, -, -)$, where $x = \{x'|x'.label = 0\}$, returns all intermediate data generated in the hidden layers, channels, and output, for input with a class label 0, in all epochs, as shown in \autoref{fig:xlabel0}. Then $H(dCNN(x, -, -, -))$ measures the entropy of this output set. Of course it may not be so helpful to calculate $H(dCNN(x, -, -, -))$ across so many layers and channels in a CNN and hence we need to further constrain the query.

\item {\bf Slice by Input, Layer, and/or Epoch:} This can be denoted as $dCNN(x, l, -, -)$ and/or $dCNN(x, l, -, t)$. When slicing on both $X$ and $L$ dimensions, we have $dCNN(x, l, -, -)$, where $l \in [0, |L|]$ specifies which layer we are interested in for a given input $x$, and $|L|$ is the number of layers in the CNN model. $dCNN(x, l, -, -)$ is the collection of all feature maps at layer $l$ for all epochs given an input $x$. To track the training process across different epochs, we need to further index the data on the $T$ dimension. For example, the entropy $H(dCNN(x, l, -, t))$ measures the entropy of output from layer $l$ during training epoch $t$ given the input $x$. Also, as an example $dCNN(X, 0, -, 0)$ is the initial condition where the input is the entire set $X$ which  not gone through any layer of the model, i.e., stays in the input layer 0.  So the layer index and the training epoch are all zeros. In this case, $H(dCNN(X, 0, -, 0))$ measures the entropy, i.e., the diversity of the input set $X$. Besides entropy, it is also possible to calculate the conditional entropy $H(a|b)$, where $a = H(dCNN(x, l_1, -, t))$ and $b = dCNN(x, l_2, -, t))$.
It describes given the output of layer $l_2$, how much we know about the output of layer $l_1$ at training epoch $t$. The channel capacity $H(a) - H(a|b) $, as defined in \autoref{eq:CC}, shows how much information gets lost between layer $l_1$ and layer $l_2$ at training epoch $t$. This can fulfill our goal of querying the information change between layers. 

\item {\bf Slice by Input, Layer, Channel, and/or Epoch:} this is denoted as 
$dCNN(x, l, c, -)$ and/or $dCNN(x, l, c, t)$. Previously, we have $H(dCNN(x, l, -, -))$ and $H(dCNN(x, l, -, t))$ measure the amount of information that a layer contains. However not every channel in this layer is useful. So we can slice the four dimensional array along the $C$ dimension to investigate a particular channel. ${dCNN(x, l, c, t)}$ returns the output (feature map) of channel $c$ in layer $l$ at training epoch $t$ given input $x$. $c \in [0, |C_{l}|]$ specifies which channel we are focused on, and $|C_{l}|$ is the number of channels in layer $l$. The entropy ${H(dCNN(x, l, c, t))}$ measures the information content of the output in channel $c$.
\end{itemize}

% First, we can treat a CNN model as a blackbox. The transmitter is the input layer while the receiver is the output layer. The communication channel is the model. Or actually, any layer inside the CNN architecture can be taken as the communication channel, with all the former layers as the transmitter and all the latter as the receiver. After this analogy, we employ information-theoretic methods to quantitatively evaluate the ability of the model. We have the following questions and corresponding experiments to verify our hypothesis. 

\subsection{Entropy Calculation for dCNN} \label{EntropyCal}
In this work, we adopt entropy as a measure of information content for CNNs. The calculation of entroy, however, needs to be specially tailored to the need of analysis. Specifically, for the four dimensional data model mentioned above, the calculation of entropy depends on how the array is sliced. We propose two types of entropy calculation for our data model. The first entropy that can be calculated is called inter-sample entropy, which is to calculate how diverse the data are in their respective high dimensional space, for example, the entropy of the training image set for classification where each image is a sample from  a high dimensional space. This type of entropy is often an indicator whether the data from the input or in the immediate layers of CNNs is sufficiently diverse,  and how the information represented in a set of samples are distilled across the different layers of a CNN. To calculate this type of entropy, we need to consider the distribution of the data samples in their embedding high-dimensional space. The second type of entropy is intra-sample entropy. The intra-sample entropy considers the entropy within each data sample. For example, the entropy of an input image or the entropy of a particular feature map. To calculate the intra-sample entropy, we use histogram-based entropy estimation. In this case, a data sample (e.g. an image) is not considered as a high dimensional point but a one-dimensional histogram by the vector component values (e.g. pixel values)  where the entropy can be efficiently calculated. 
Below we describe the calculation methods in detail. 

\subsubsection{Inter-Sample Entropy Calculation}
% For example, Given a data model $CNN(x, -, -, -)$, when $x$ is the set of all training images in MNIST dataset. 
The purpose of inter-sample entropy calculation is to measure the diversity of the data samples in their embedding space, which is often  high-dimensional. The distribution of data samples in the high-dimensional space is the basis for classification, where the task is to identify the decision boundaries between data of different classes. As the set of samples are going through the different layers of a neural network, redundant or irrelevant information for the inference task is discarded which will in turn change the distribution of the layer output. Monitoring how the entropy is changed often provides an important hint on how the neural network is doing. As an example, considering our $dCNN$ data model and assuming $X$ is the set of {\em all} training images in the MNIST dataset where each data sample in $X$ is an image of size 28*28 in a 784-dimensional space. If we want to measure the diversity of the training set  $X$, we can first slice the CNN data array by $dCNN(X, 0, -, 0)$, and then calculate the inter-sample entropy, denoted as $H(dCNN(x, 0, -, 0))$, to measure whether the input samples have sufficient diversity. Below we explain how the inter-sample entropy is computed. 

% \begin{figure}[htp]
%     \centering
%     \includegraphics[width=8cm]{pictures/pij.pdf}
%     \vspace{-9pt}
%     \caption{Convert distances into probability distributions. Distances between point a and other points are converted into }
%     \label{fig:pij}
% \end{figure}

As defined in \autoref{eq:Entropy}, when calculating the entropy, we need the probability for each of the states in the system. To obtain the probabilities for calculating the inter-sample entropy, we adopt an idea inspired by an dimensionality reduction technique, Uniform Manifold Approximation and Projection (UMAP) \cite{McInnes:2018:UMAP}. In UMAP, to create a distribution for high dimensional points, the distance between a pair of samples $i$ and $j$ is converted into a probability by an exponential distribution function. With the probabilities between all sample pairs, we can then calculate the inter-sample entropy as the expected value of the information content for all sample pairs. 

The exponential distribution is widely used to model relationships between random variables. For any data point $x_i$, the similarity between $x_i$ and another point $x_j$ is given by the conditional probability $P_{j|i}$:
%We have two approaches to compute the  probabilities. The first uses the Gaussian distributions to convert Euclidean distances in high-dimensional space into probabilities, and the second utilizes exponential distributions as the mapping function.

%\vspace{-4pt}
%\begin{itemize}
% \item Gaussian Distribution Based Probability Calculation: \\
%   Similar to the first step of Stochastic Neighbor Embedding \cite{Hinton:2018:SNE}, the similarities between data points in high-dimensional space can be quantified by Gaussian distributions. 
%   To be specific, choose a data point $x_i$ in the high-dimensional space, the probability of choosing another point $x_j$ is defined as \cite{Hinton:2018:SNE}:
%   \begin{equation} \label{eq:p_j_given_i_Gaussian}
%   P_{j|i} = \frac{exp(-\|x_i-x_j\|^2 / 2 \sigma_i^2)}{\sum_{k\neq i}exp(-\|x_i-x_k\|^2 / 2 \sigma_i^2)}
%   \end{equation}
  
%   \noindent where $\sigma_i$ is either set by user or found by a binary search to fulfill a ``perplexity'' \cite{Maaten:2008:tsne} constraint defined by user.
%   $P_{j|i}$ measures the similarity of data point $x_j$ relative to data point $x_i$. The similarity is defined proportion to the probability density function of Gaussian with mean $x_i$ and and variance $\sigma_i$. If the point $x_j$ is far from $x_i$, then $P_{j|i}$ is smaller. In our implementation, we use a constant user-defined $\sigma$ for every data point $x_i$. A larger $\sigma$ is suitable for modeling sparse data distribution and a small one is better for dense distribution. 
%   \vspace{-4pt}

  \begin{equation} \label{eq:p_j_given_i_Exp}
  P_{j|i} = exp(\frac{-\|x_i-x_j\|}{\sigma_i}) 
  \end{equation}
  % not necessarily Euclidean distances like tSNE but rather any distance can be plugged in
  where $\sigma_i$ is the scale parameter of the distribution for each $x_i$. An adaptive exponential kernel for each point are more powerful to model the real data distribution in high-dimensional space, as a result, each $\sigma_i$ is set to satisfy \autoref{eq:sigma_i}:
  \begin{equation} \label{eq:sigma_i}
  \sum_{j=1}^k exp(\frac{-\|x_i-x_j\|}{\sigma_i}) = log_2(k)
  \end{equation}
  where $k$ is the number of data points. 
%   Compare to Equation \ref{eq:p_j_given_i_Gaussian}, we do not have a normalization step in Equation \ref{eq:p_j_given_i_Exp}, but the probabilities are set so that they add up to a constant value. From a computational view, this omition can reduce the computational time since summation is a computational expensive operation and has non negligible impact on the performance.
% \end{itemize}
%\vspace{-4pt}

To make the above calculation computationally efficient, for each sample, we only consider its k-nearest neighbours in the high-dimensional space when performing the probability calculation. After the calculation, the probabilities are symmetrized and normalized before computing the entropy. To make them symmetric, we average $P_{j|i}$ and $P_{i|j}$: 
\begin{equation} \label{eq:pij}
P_{i,j} = \frac{P_{j|i} + P_{i|j}}{2N}
\end{equation}
where $N$ is the number of data points. Then we normalize the joint probability table. Now we have probability for each of the states, we plug \autoref{eq:pij} into \autoref{eq:Entropy} for inter-sample entropy calculation:
\begin{equation} \label{eq:inter_Entropy}
H(X) = -\sum_{i=1}^n\sum_{j=1}^n P_{i,j}log_2P_{i,j}
\end{equation}

\subsubsection{Intra-Sample Entropy Calculation} \label{Intra-Entropy}
The intra-sample entropy measures the randomness of the values in each data sample, regardless of the data sample's embedding dimensionality. For example, when estimating the entropy for an image or a feature map, we only takes the value distributions of the pixels into consideration. In this case, the most commonly used probability estimation approach is based on the histogram. This approach is often adopted to measure the quality of images for image processing applications.  

The calculation details are as follows. First, we normalize the value range to $(-1, 1)$, divide the range into $B (B=32$ for example) adjacent equal-size intervals (bins), and then put the values of a data sample (e.g. an image's or a feature map's pixel values)  into these bins. The resulting frequency distribution is a histogram. Then we use the normalized frequency distribution as the probability distribution to calculate the intra-sample entropy. Although straightforward, this approach is computationally efficient and still can measure the sharpness or blurriness of value distributions. 

One example of using intra-sample entropy is when we want to calculate the randomness within a group of feature maps, which is denoted as $H(dCNN(x, l, c, t))$ where $x$ is a set of input images from the same class. This is useful for the evaluation and understanding of CNN's filters. Because a `dead' filter will not be activated by different inputs, the value distribution on one channel can give some idea about the corresponding filter. It is feasible that we use intra-sample entropy for each feature map and use the averaged result as the final $H(dCNN(x, l, c, t))$. 

\section{Information Query via CNNSlicer} 
In the previous section, we formulate a 4-dimensional hypercube $dCNN(X, L, C, T)$ to represent a CNN model. By slicing on the hypercube we can extract information related to the CNN for entropy-based analysis. 
To meet our requirements stated in Section \ref{requirement}, we develop a visual analysis system, called {\em CNNSlicer},  with four visual components. In this Section, we describe the visual components of CNNSlicer and show how various entropy measures can assist the analysis of CNN models. 

% As illustrated in \autoref{fig:datamodelView}(A), when only slicing on the $X$ (input) dimension on the data model, we will get a collection of all data samples related to the input, including intermediate results and the output for all epochs. \autoref{fig:datamodelView}(B) shows the corresponding slices at epoch $t$ in the hypercube when users select two layers $l_i$ and $l_j$ in the CNN architecture View. 

\subsection{Query Selection and Visualization}
There are various of information queries users can make according to different slicing operations. To assist users to make proper information queries, we design a query view which allows users to form information query in two ways. 

\begin{figure}[htp]
    \flushleft
    \includegraphics[width=9.5cm]{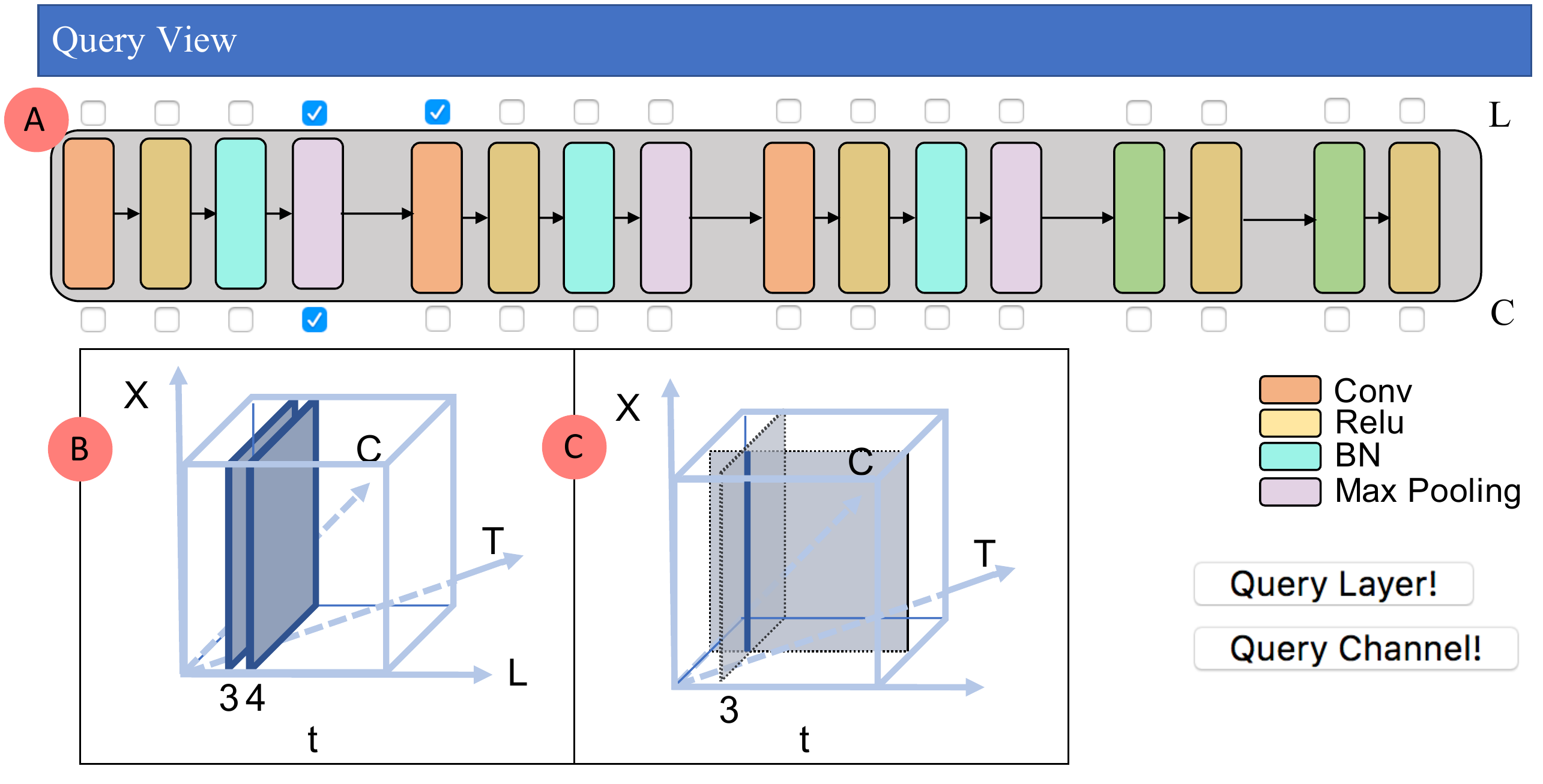}
    \vspace{-9pt}
    \caption{Query view. (A) overview of the CNN architecture for users to make selection of layers; (B) Selecting layers using CNNSlicer; (C) Selecting a channel by the intersection of layer and channel slices. }
    \label{fig:queryView}
\end{figure}

The first is through an overview of the CNN architecture as shown in \autoref{fig:queryView}(A). With all layers visually available in the CNN architecture, the slicing on the layer dimension becomes more straightforward. Users can make selections of layers in this view for further analysis to support the top-down analysis \textbf{(R1)} of the CNN model. Another way is though a visualization of the data model s shown in \autoref{fig:queryView}(B) and \autoref{fig:queryView}(C). To have a conceptual understanding of information queries, it can be very helpful to visualize the slicing operations on the data hypercube $dCNN(X, L, C, T)$ directly \textbf{(R2)}. In \autoref{fig:queryView}(B), Users can drag to select layers they are interested in. By dragging on both layer and channel dimensions in \autoref{fig:queryView}(C), users can select a specific channel to analyze. After the selection, users need to click the ``Query'' button to update other visual components.

 \vspace{-4pt}
 
\subsection{Evaluate Input and Output} 
\begin{figure}[htp]
    \centering
    \includegraphics[width=9cm]{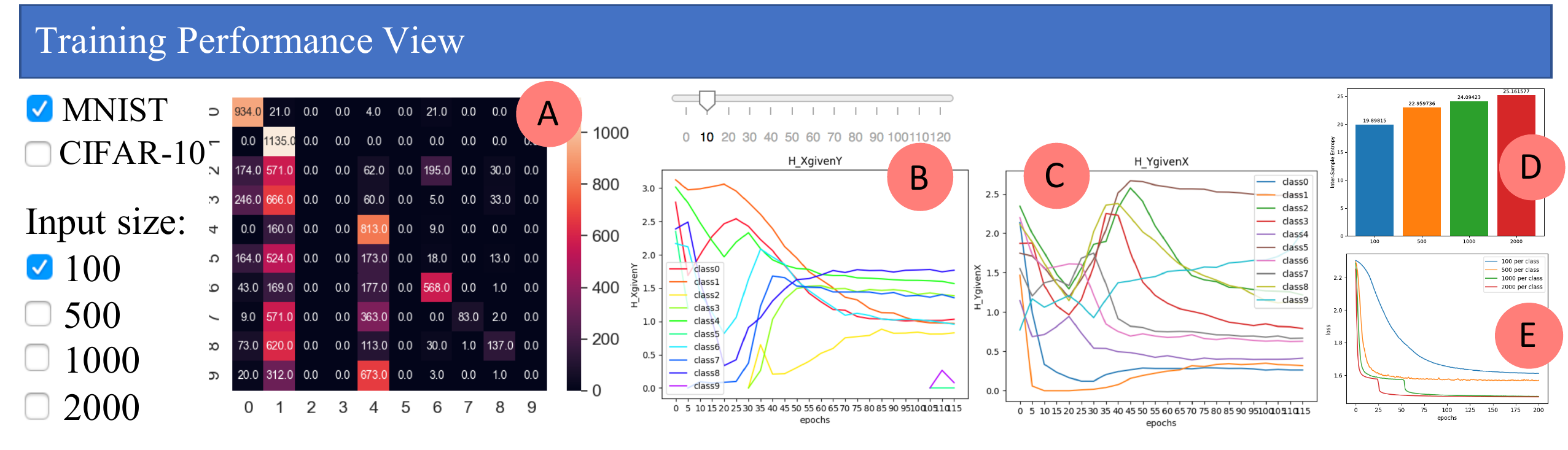}
    \vspace{-12pt}
    \caption{Training performance view. (A) confusion matrix; (B) entropies of input given output; (C) entropies of output given input; (D) inter-sample entropies; (E) training loss.}
    \label{fig:TrainPerform}
\end{figure}

As discussed before, $dCNN(x, 0, -, 0)$ is a special case of slicing on the hypercube, which returns the input $x$ to the model.  Given the input $x$, output of the CNN model at training epoch $t$ can be denoted as $dCNN(x, |L|, -, t)$, where $|L|$ is the number of layers in the model. 
If we slice $dCNN$ on the input and output layers, answers to the information queries about them  \textbf{(R3)} can help users evaluate the training performance of the model. As a result, we design a training performance view, as shown in \autoref{fig:TrainPerform}, to show the related statistics. In this section, we present the entropy calculation and related visualizations in detail. 

\begin{itemize} 
    \item \textbf{Information queries about the input.}
    Increasing the training data size will usually improve the model's robustness and accuracy. However, if the training data is not well distributed, it can lead the model to optimize in a nonoptimal direction and affect the speed of convergence. Therefore, it would be worth to evaluate the diversity of the input. To do this, we use inter-sample entropy $H(dCNN(x, 0, -, 0))$ to measure the diversity of data samples in high-dimensional space. 
    
     \begin{figure}[htp]
    \centering
    \includegraphics[width=8.5cm]{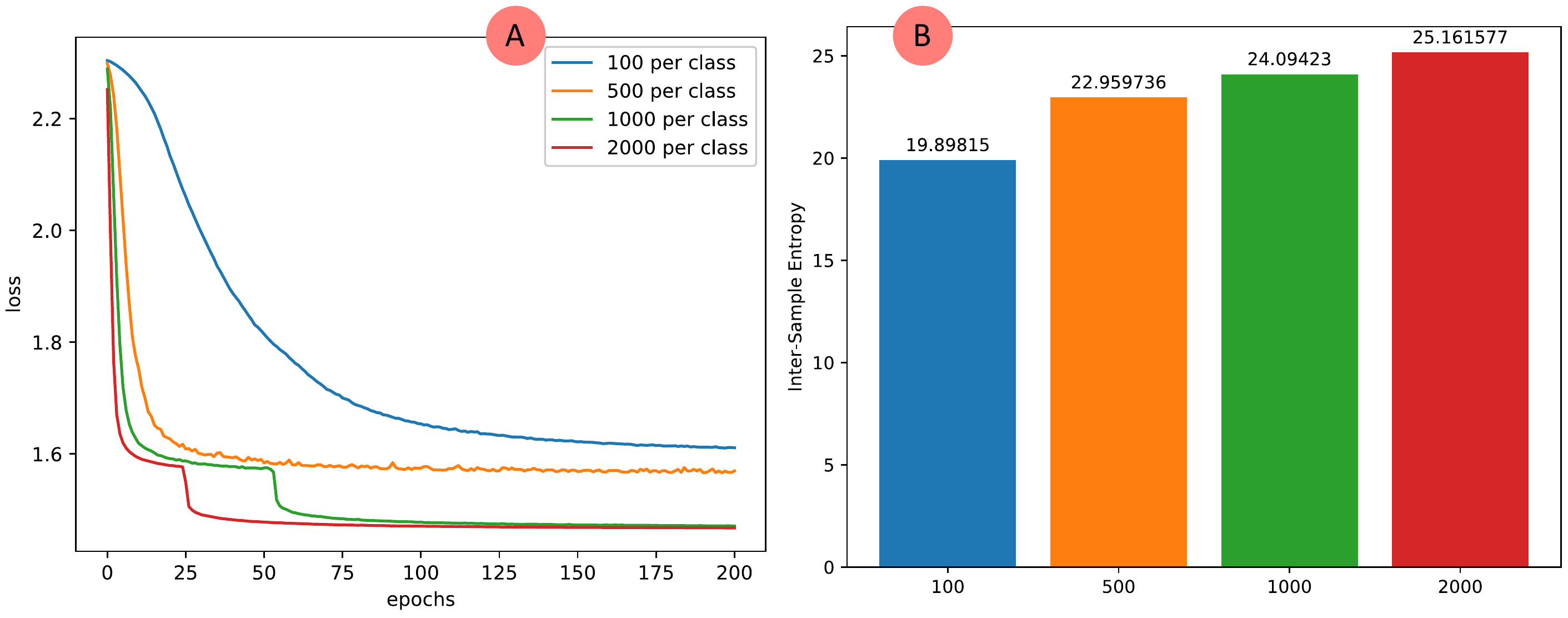}
    \vspace{-12pt}
    \caption{(A) Training loss for datasets with 100, 500, 1000, and 2000 samples per class; (B) inter-sample entropy for datasets with 100, 500, 1000, and 2000 samples per class.}
    \label{fig:loss}
\end{figure}

    Take MNIST dataset as an example. This dataset has ten classes. First, we random sample 100, 500, 1000, and 2000 samples respectively from each class resulting in four datasets ($x1, x2, x3, x4$) with different sizes. To analysis the diversity of the samples, we calculate the inter-sample entropy $H(dCNN(x_i, 0, -, 0))$ with $i \in [1, 4]$. As shown in \autoref{fig:loss}(B), the horizontal axis shows the number of samples per class and the vertical axis is the inter-sample entropy for the dataset. We retrain the model based on these datasets separately. The training performance (loss) is shown in \autoref{fig:loss}(A) where the horizontal axis represents training epoch and the vertical axis is the loss value. \autoref{fig:loss}(A) and \autoref{fig:loss} (B) share the same color scheme which makes it easier for users to compare different datasets and connect the training performance with the diversity of training data. For example, blue represents the dataset with 100 samples per class. We find the blue line in \autoref{fig:loss}(A), compared to the other lines,  takes more that 150 epochs before the loss becomes flat and stable. The blue bar in \autoref{fig:loss}(B) shows this dataset has relative low inter-sample entropy (i.e., 19.89815) compared to other datasets (i.e., 22.959736, 24.094230 and 25.161577), meaning the data is less diverse. On the other hand, the dataset with 2000 samples per class has the highest diversity, and the training loss is more steep and converges much faster (at about epoch 50). 
    
    Thus, we have the conclusion that dataset with high inter-sample entropy, i.e.,  diversity, do converge faster. 
    % which is consistent with the common knowledge.
    
    % {\color{red}{For every dataset, we compute the inter-sample entropy $H(dCNN(x, 0, -, 0))$ for each class to measure the diversity inside a class. In this case, $x = \{x'|x'.label = i\}$ and $i \in [0, 9]$. A stacked bar chart is adopted to show these inter-sample entropies in Figure \ref{fig:TrainPerform}(D). These plots give users an overview of the model's performance. }} 
    % Our hypothesis is that for some classes with high-variance (high entropy), they will have better performance during testing since the training data for this class has more variations. 

    \item \textbf{Information queries about the training performance.} Think of the classification process of a CNN model as a communication channel with the input layer transmitting the input $x$ to the output layer through the model. We can make information queries about input $x$ and output $y = dCNN(x, |L|, -, t)$.

    As discussed in Section \ref{information_theory}, $H(x.label|y=i)$ measures given the predicted labels $y$ being class $i$, how much uncertainty we have about the ground truth labels $x.label$. On the other hand, $H(y|x.label=i)$ measures the randomness of the model’s predictions in the output for the data samples in class $i$ at epoch $t$. The calculation of $H(x.label|y=i)$ and $H(y|x.label=i)$ can be done via the classic Shannon's entropy definition. That is, for a given class $i$, to calculate $H(y|x.label=i)$ we need $p(y=j|x.label=i)$ where $j \in [0, 9]$. These probabilities can be otained from the model output. Then we can directly use the Shannon's entropy equation to evaluate. 
    
    To clearly visualize the training performance of the CNN classification model, we have a confusion matrix panel and two plots for $H(x.label|y=i)$ and $H(y|x.label=i)$ as shown in \autoref{fig:TrainPerform}(A), \autoref{fig:TrainPerform}(B) and \autoref{fig:TrainPerform}(C). The confusion matrix shows how confused the model is between two classes. Each row of the confusion matrix represents all instances of predicting an actual class to all other classes, including the correct class. Each column represents all instances of different actual classes predicted as this class. The cell color indicates how many instances are predicted as the column class but actually belongs to the row class. The higher value the darker is the color. In a line chart, the horizontal axis is the training epoch and the vertical axis is the entropy value. Thus, these line charts can assist users to monitor the entropy change over training epochs.

    We use the MNIST dataset as an example to demonstrate the usefulness of our visual design. Users can select different input sizes to evaluate. After the selection, the corresponding confusion matrix and line charts get updated. \autoref{fig:HXgivenY} shows a comparison of $H(x.label|y=i)$ between datasets with input size 1000 (\autoref{fig:HXgivenY}(A)) and 10000 (\autoref{fig:HXgivenY}(B)) respectively. We find $H(x.label|y=i)$ gets stabilized faster (at about epoch 25) with input size 10000. But $H(x.label|y=i)$ still dramatically goes ups and downs until epoch 70 with input size 1000. We conclude that training with larger input data gets stabilized faster, and ``stabilized'' here means the uncertainty about input class given output gets stable. 
    \begin{figure}[htp]
    \centering
    \includegraphics[width=9cm]{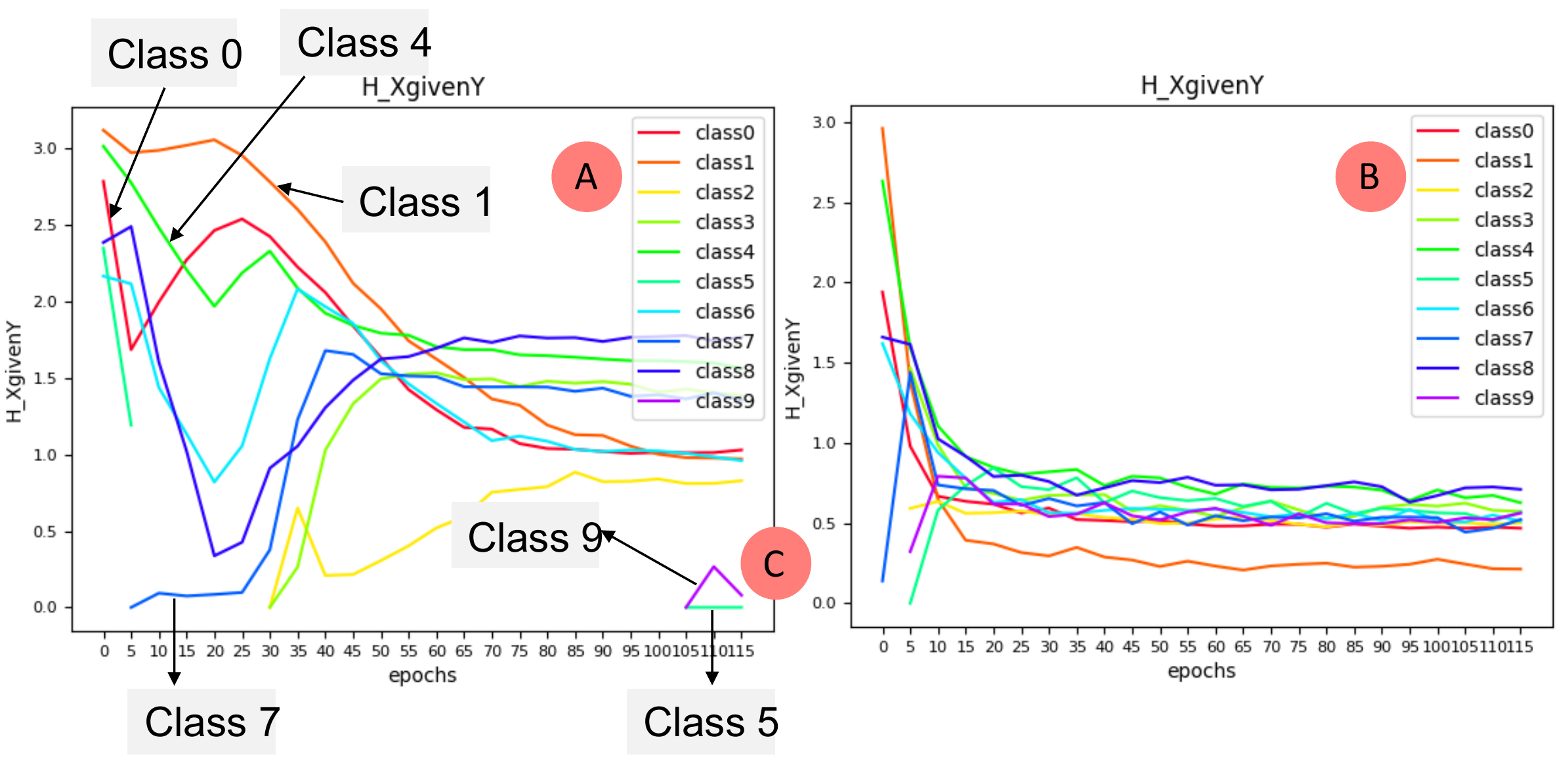}
    \vspace{-12pt}
    \caption{(A) $H(x.label|y=i)$ with input size 1000,  $i \in [0, 9]$; (B) $H(x.label|y=i)$ with input size 10000, $i \in [0, 9]$. }
    \label{fig:HXgivenY}
    \end{figure}
    
     \vspace{-9pt}
     
    \begin{figure}[htp]
    \centering
    \includegraphics[width=7cm]{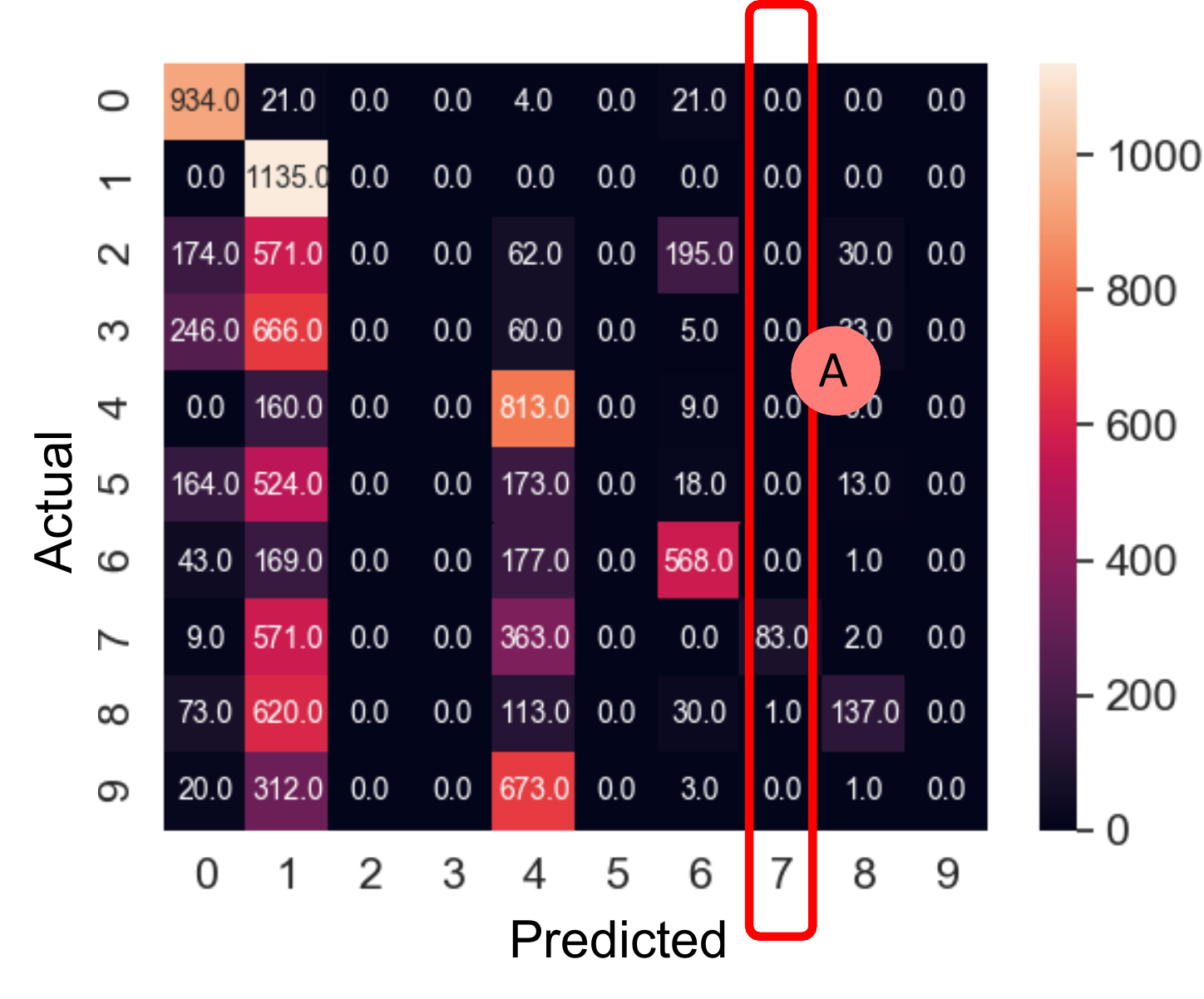}
    \vspace{-12pt}
    \caption{Confusion matrix for training epocch 10.} \label{fig:heatmap}
    \end{figure}
    
    In \autoref{fig:HXgivenY}(A), we notice that at the beginning, $H(x.label|y=0)$, $H(x.label|y=1)$ and $H(x.label|y=4)$ are high which indicates some instances belong to other classes get wrongly classified into class 0, 1 and 4. In \autoref{fig:heatmap}, the brighter columns (column 0, 1 and 4) in confusion matrix verifies this. In \autoref{fig:HXgivenY}(A), $H(x.label|y=7)$ is low at the beginning epochs. This means the model does not have too much randomness during prediction for this class. The low entropy here is a reflection of high precision score. As shown in \autoref{fig:heatmap}(A), among all instances classified into class 7, most of them are actually belong to class 7. The segments in \autoref{fig:HXgivenY}(C) shows $H(x.label|y=5)$ and $H(x.label|y=9)$ are infinity in previous epochs. In \autoref{fig:heatmap}, we can see column 5 and column 9 are all zeros meaning no instances get classified into these two classes. The model has a hard time learning to predict class 9 and class 5. This is the information we can not get from an overall accuracy score. 
    
    % Figure \ref{fig:TrainPerform}(C) describes given the actual input label, how does the entropy of predictions change over the course of training. Class 9 and class 5 have relative high entropies in later entropies indicating these instances are incorrectly classified into other classes. 
    
    From several case studies, we shows that it is effective to use entropy as the evaluation metric when visualizing the training performance. The visualization clearly tells us when the training gets stabilized. Besides, unlike the accuracy measure which only describes how much instances get correctly classified, entropy also encodes information about wrong predictions such as how uncertain (random) the predictions are. Entropy ($H(x.label|y=i)$, $i \in [0, 9]$) can help users locate when the model has high precision. However, low entropy does not indicate good performance since all samples can be wrongly classified into one class and still has the low entropy, although the chance for this to happen is not high. . 
\end{itemize}

% \subsection{Visualize Slicing Operations on dCNN}
% To have a systematic view of queries and also understand all possibilities when exploring a CNN model, it is necessary to visualize the slicing operations on the data model $dCNN(X, L, C, T)$. As a result, we have a data model view \textbf{(R2)} showing the current slicing operation. 
% \begin{figure}[htp]
%     \centering
%     \includegraphics[width=7cm]{pictures/datamodelView.pdf}
%     \vspace{-9pt}
%     \caption{(A) Data model view; (B) slice by input (X), layer (L), and Epoch (T) dimensions.}
%     \label{fig:datamodelView}
% \end{figure}

% As illustrated in \autoref{fig:datamodelView}(A), when only slicing on the $X$ (input) dimension on the data model, we will get a collection of all data samples related to the input, including intermediate results and the output for all epochs. \autoref{fig:datamodelView}(B) shows the corresponding slices at epoch $t$ in the hypercube when users select two layers $l_i$ and $l_j$ in the CNN architecture View. 

\subsection{Evaluate the Information Between Layers}
In the previous sections, we have discussed information queries about input ($dCNN(x, 0, -, 0)$) and output ($dCNN(x, |L|, -, t)$) which are two special cases of $dCNN(x, l, -, t)$. In this section, we evaluate more general case which is the information flow between the intermediate layers. 
To visualize the amount of information transmitted between layers, we design a layer view to show the information statistics (i.e., inter-sample entropy) \textbf{(R1, R3)}. With the help of the CNN architecture view, users can easily slice on the layer dimension of the hypercube. Also, the layer view contains visualization of feature maps \textbf{(R4)} to help users get insight of the layer. Details of information statistics and visual analysis are as follows.

Suppose we are interested in how the information flows from layer $l_i$ to layer $l_j$ during training. Then all layers before $l_i$ can be taken as a transmitter and all layers after $l_j$ can be taken as a receiver. The stacked layers between layer $l_i$ and $l_j$ represent the noisy communication channel. We use the channel capacity in \autoref{eq:CC} to measure the amount of information between layer $l_i$ and layer $l_j$ at epoch t: $Capacity = H(X) - H(X|Y)$ where $X = dCNN(x, l_i, -, t)$ and $Y = dCNN(x, l_j, -, t)$. 

\subsubsection{ Information between two channels} Normally, each layer's output has multiple channels generated by different filters. To reduce the analysis complexity and make the explanation process more clear, we consider each channel as one representation of this layer's output. That is, given a group of input instances, the channel capacity between layer $l_i$ and layer $l_j$ is calculated based on the feature maps from all channel pairs, one channel from layer $l_i$ and another from layer $l_j$. However, feature maps from layer $l_i$ and layer $l_j$ are in two spaces with different dimensionality. To compute their joint distributions, we adopt a similar idea that uses distances in high-dimensional space between two samples (i.e. two feature maps produced by the same channel from two different input images) as the similarity measurement. Instead of converting distances into probabilities using exponential or Gaussian distributions, we map the distances into a 2-dimensional histogram, where one axis represents the bins that discretize the pairwise feature map distances in a channel of layer $l_i$, and the other axis represents the bins of the distances in a channel of  layer $l_j$. Any entry in the 2D histogram records the frequency of the joint distance pairs from the two channels, one from layer $i$ and another from layer $j$, computed from the feature maps. 
After the 2D histogram is established, we can compute the joint entropy of two channels, the entropy of each channel and the  conditional entropy between the channel pairs by marginalization, and finally, the channel capacity between the channel in layer $i$ and the channel in layer $j$. The 2D histograms are useful in this case since it combines these two high-dimensional spaces and study the distance correlation between two filters and can be used to calculate the channel capacity. One alternative approach is to use a 2D-Gaussian distribution that take the distance pairs from two layers as input to do the probability mapping. Due to the computational complexity, we use the approach based on 2D histograms. 

\begin{figure}[htp]
    \centering
    \includegraphics[width=9cm]{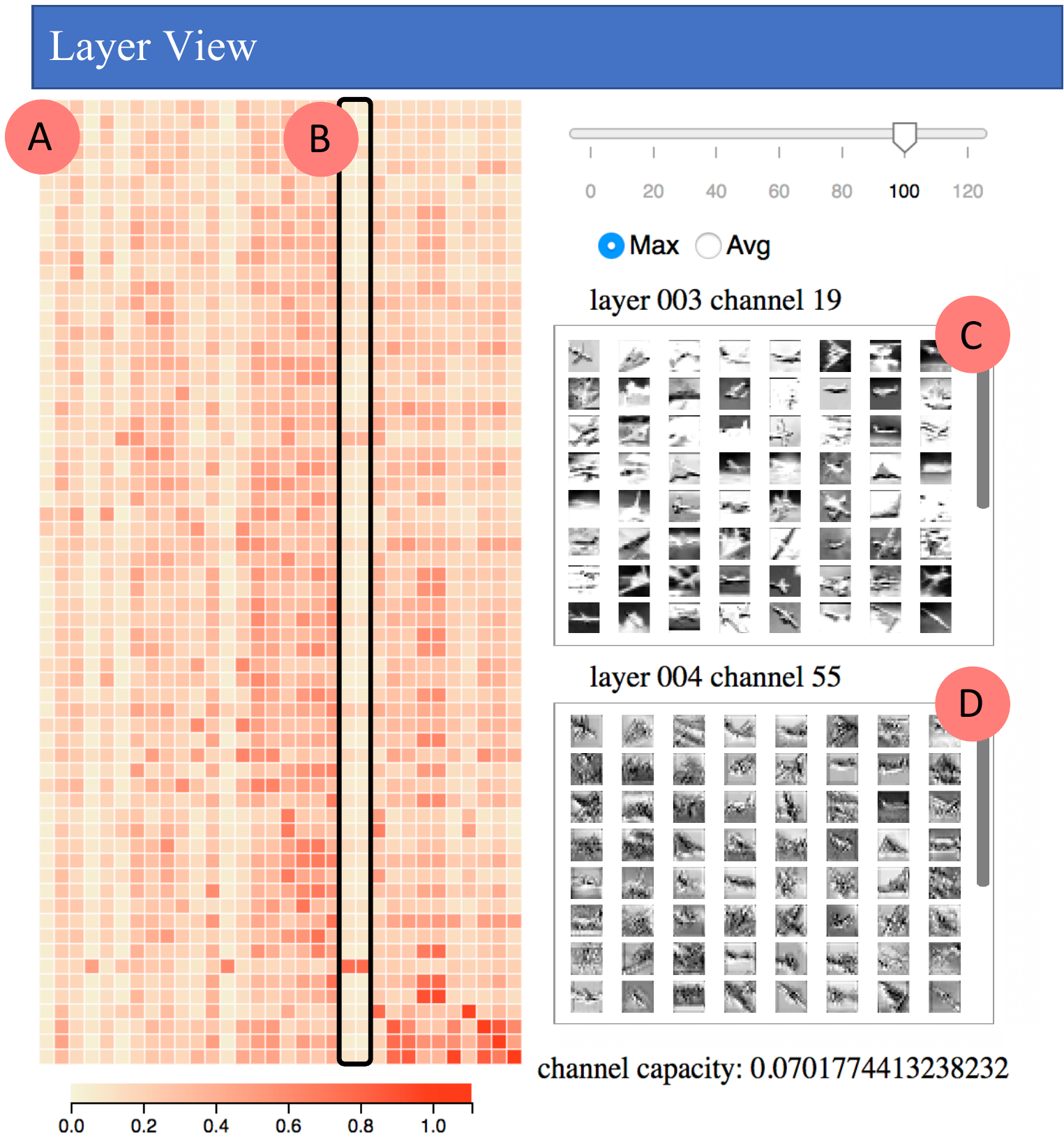}
    \vspace{-12pt}
    \caption{Layer view. (A) Heatmap of channel capacity between layer 3 (horizontal axis) and layer 4 (vertical axis); (B) two filters with similar pattern highlighted in the black rectangle; (C) feature maps from filter 19 in layer 3; (D) feature maps from filter 55 in layer 4.}
    \label{fig:layerView}
\end{figure}

\subsubsection{Information matrix between two layers} After we know how to calculate the channel capacity between any two channels, one from each layer, to get a clear insight into the information changes between two layers, we use a sorted heatmap to visualize the matrix of  channel capacity between the  channels from two layers. An example is shown in \autoref{fig:layerView}(A). The horizontal axis of this heatmap are channels of layer $l_i$ and the vertical axis are channels of layer $l_j$. Color at position $(x, y)$ in the heatmap encodes the channel capacity between layer $l_i$ and layer $l_j$ using the feature maps from the $x$-th filter of layer $l_i$ and the $y$-th filter of layer $l_j$. A brighter color indicates a smaller channel capacity between these two channels which means the channel undergoes higher information loss, or less correlation between the two channel outputs. One column of this heatmap represents the channel capacity calculated based on feature maps from one filter in layer $l_i$ and feature maps from all filter in layer $l_j$. If there are 64 filters in layer $l_j$, then there will be 64 channel capacity values in this column. Users can sort the heatmap by rows or columns based on the average or maximum value within the row or column vectors. More specifically, by sorting the matrix columns by the maximum values we meana: (1) compute the maximum value of each column; (2) sort these maximum values and keep the sorted indexes; (3) rearrange the columns based on these indexes. We can perform similar operations for rows if so desired. After sorting, patterns of channel capacity between layers become clear. When users click on one rectangle of the heatmap, feature maps from the corresponding``column filter'' and ``row filter'' will show up. \autoref{fig:layerView}(C) and \autoref{fig:layerView}(D) are the results when we click on the lower left rectangle. 

\subsubsection{Exploration of Information Matrix}
\begin{figure}[htp]
    \centering
    \includegraphics[width=7.5cm]{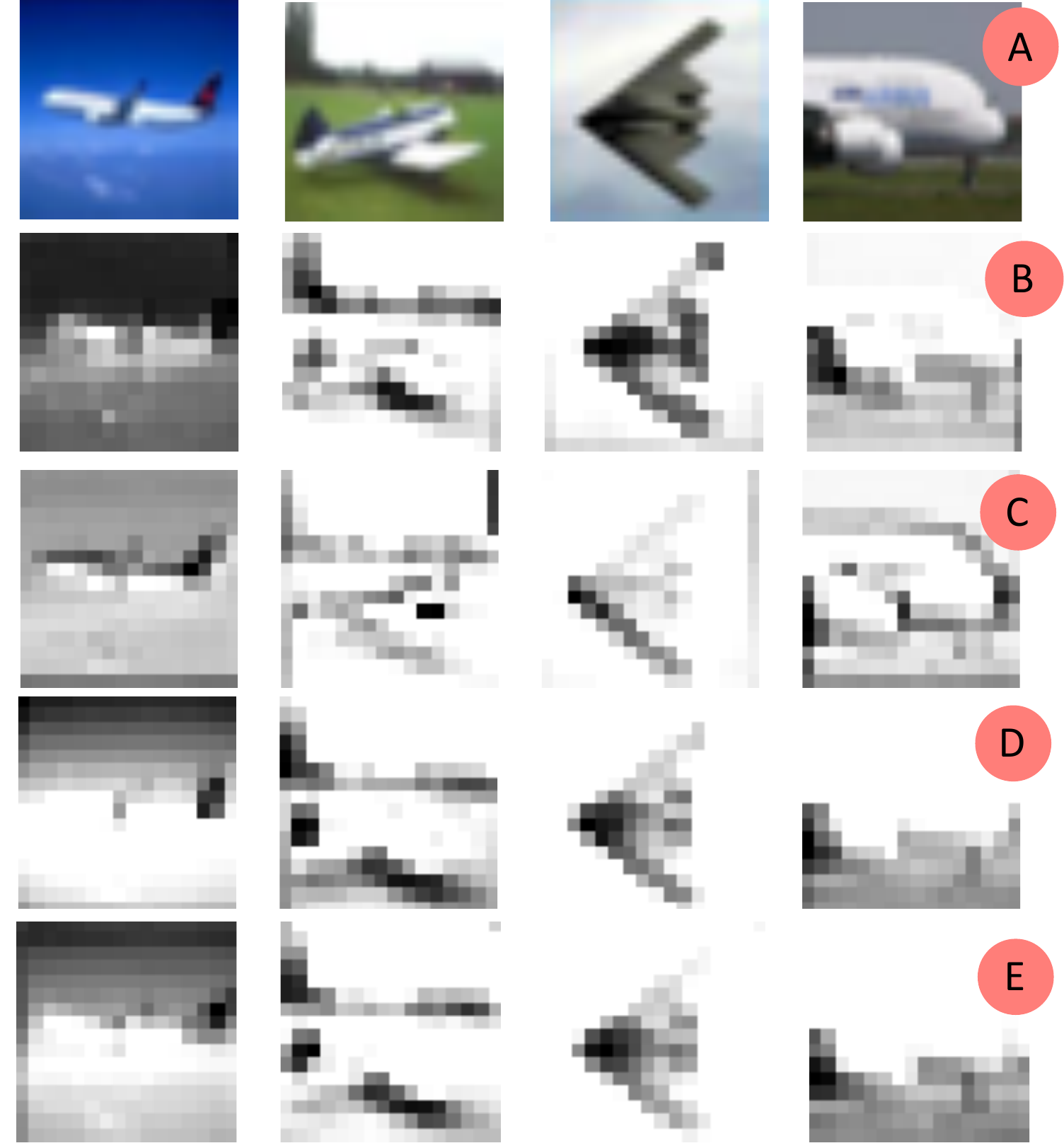}
    \vspace{-8pt}
    \caption{(A) original input; (B) to (E) are feature maps at layer 3 from filter 0, filter 24, filter 8, filter 30 respectively.}
    \label{fig:f8f30}
\end{figure}

We adopt the CNN classification model in \autoref{fig:queryView}(A) trained on the CIFAR-10 dataset for case studies. First we choose layer 3 (horizontal axis) and layer 4 (vertical axis) in the query view shown in \autoref{fig:queryView}(A). We select epoch 100 and sort the heatmap by maximum value for each dimension. In \autoref{fig:layerView}(A) we find two filters (filter 8 and filter 30 in layer 3) highlighted in the back rectangle in (B) have similar patterns. Our hypothesis is that these two filters are similar. To verify our hypothesis, we visualize the feature maps from these two filters and other two random selected filters in layer 3 as shown in \autoref{fig:f8f30}. \autoref{fig:f8f30}(A) are the original input images. In \autoref{fig:f8f30}, from the second row to the bottom row, they are feature maps from filter 0, filter 24, filter 8 and filter 30 respectively. We can see filter 8 and filter 30 perform similarly (\autoref{fig:f8f30}(D) and \autoref{fig:f8f30}(E)) which verifies our hypothesis.  

% After sorting this heatmap by average value in each dimension, we pick the first row of this feature map which representing filter 52 in layer 4. According to our definition of channel capability, this filter cause high information loss between layer 3 and layer 4. The blurry feature maps of this layer in Figure XX(A) shows this filter do lose some essential information (such as boundary of input) compared to feature maps from other filters in Figure XX. 

\begin{figure}[htp]
    \centering
    \includegraphics[width=8cm]{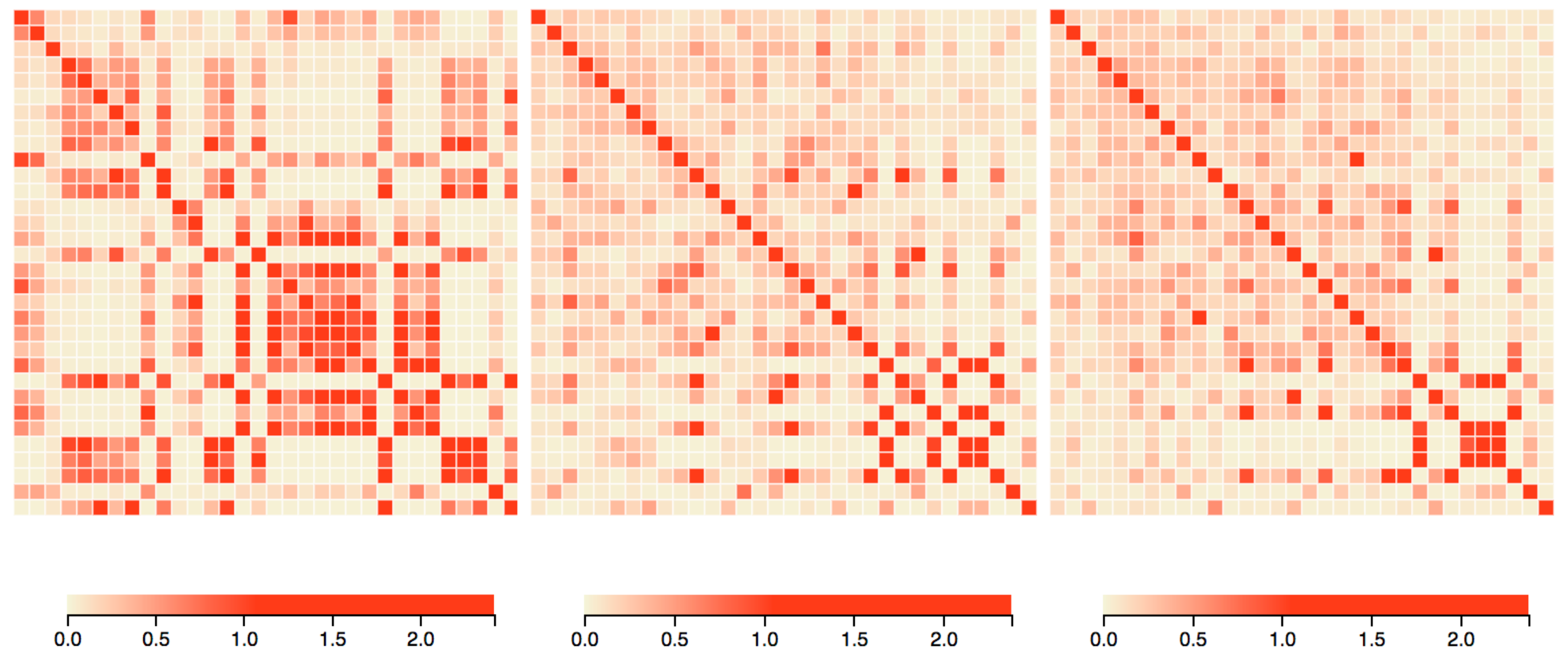}
    \vspace{-11pt}
    \caption{From left to right, they are heatmaps of channel capacity between layer 2 and layer 3 sorted by maximum value in each dimension at epoch 0, epoch 40 and epoch 120.}
    \label{fig:l2l3_0_40_120}
\end{figure}

We are also interested in the information flow between layer 2 and layer 3. The control of information flow here is mainly done by max-pooling. When we drag the epoch scrollbar, we can see the change of channel capacity over the training time in \autoref{fig:l2l3_0_40_120}. From left to right, they are heatmaps of channel capacity at epoch 0, 40, and 120 respectively. Because we are investigating the information change before and after an max-pooling, it makes sense that the diagonal of these heatmaps have highest channel capacity. In the earlier training epochs, there are lots of rows and columns having similar patterns, meaning some filters are redundant. Through iterative training, there are less dark square patterns and similar rows and columns are decreasing which means filters in the convolutional layer before this activation are learning to extract useful features. 

\begin{figure}[htp]
    \centering
    \includegraphics[width=9cm]{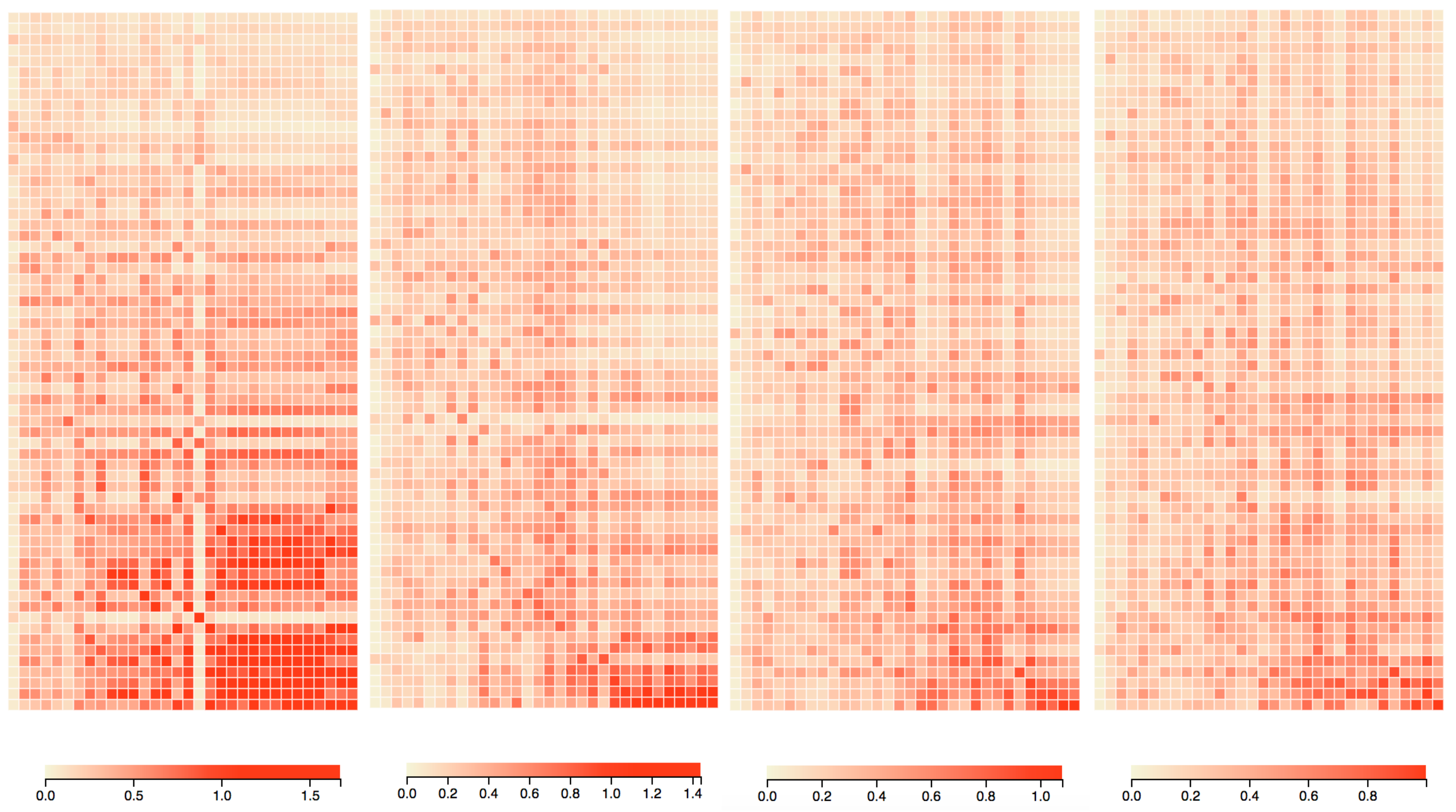}
    \vspace{-11pt}
    \caption{From left to right: heatmaps of channel capacity between layer 1 and layer 5 sorted by maximum value in each dimension at epoch 0, epoch 20, epoch 80 and epoch 120 respectively.}
    \label{fig:l1l5_0_20_80_120}
\end{figure}

\begin{figure}
    \centering
    \includegraphics[width=9cm]{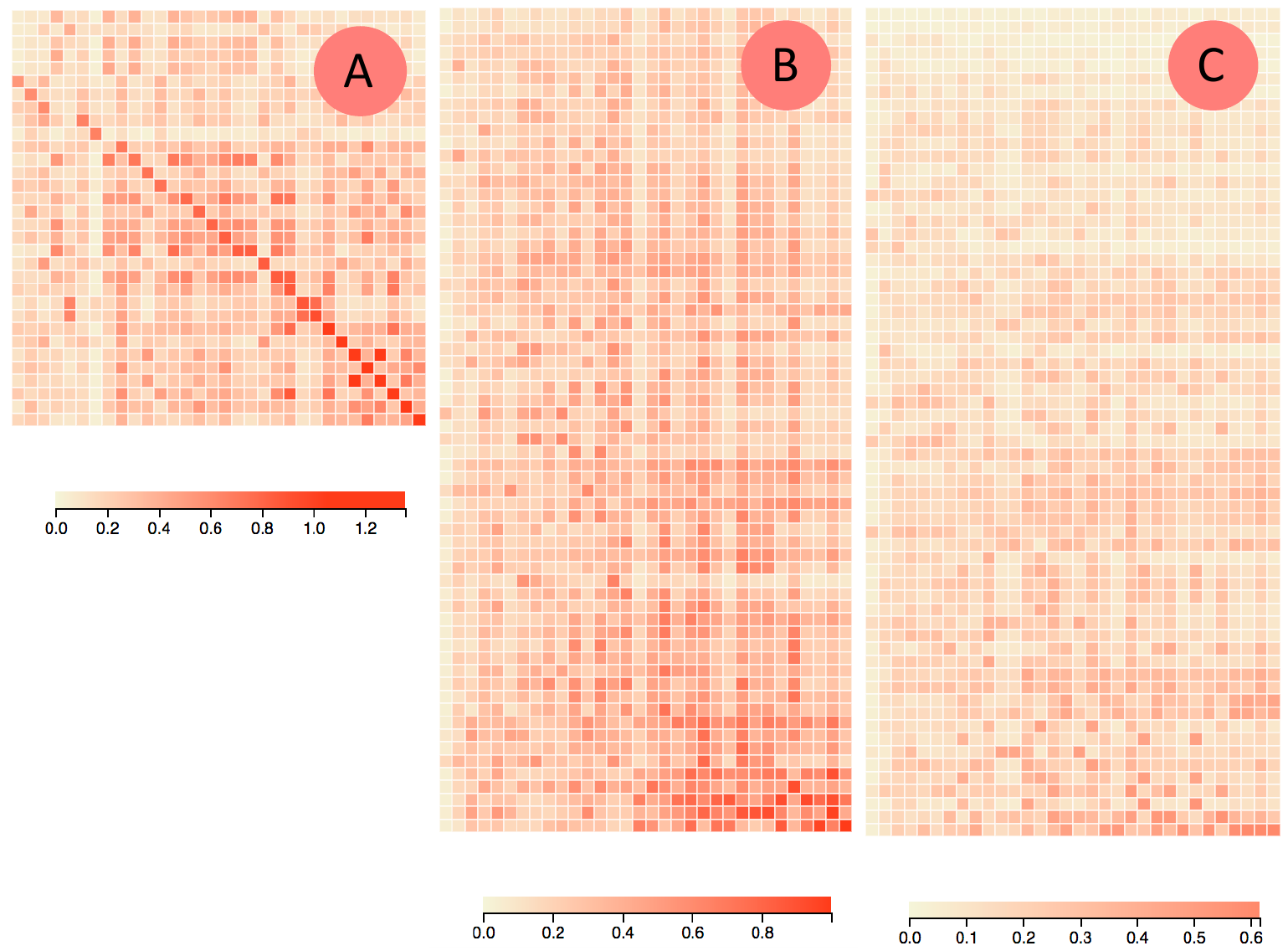}
    \vspace{-11pt}
    \caption{(A) Channel capacity between layer 1 and layer 3; (B) channel capacity between layer 1 and layer 5; (C) channel capacity between layer 1 and layer 7.}
    \label{fig:l1_357}
\end{figure}

We also compare the channel capacity between layer 1 and layer 5 which are the activations of two convolutional layers. \autoref{fig:l1l5_0_20_80_120} shows heatmaps of channel capacity between these two layers at epoch 0, 20, 80, and 120 respectively. There is an obvious trend that the heatmap is getting brighter (smaller capacity values) and the maximum value of each heatmap is getting lower which means the model is trained to distill the input information and make feature maps tight in high-dimensional space. 

To investigate the connection between the formation flow and the depth of layers, we compare the channel capacity between layer 1 and some latter layers (i.e., layer 3, layer 5, layer 7) at epoch 120. Intuitively, with each layer of the neural network learning to reduce the irrelevant information from input, the channel capacity between the input and deeper layers decreases. This explains the phenomena that at epoch 120, the heatmaps of channel capacity between these layers are getting brighter as we approach to deeper layers, as shown in \autoref{fig:l1_357}, where (A), (B) and (C) are heatmaps of channel capacity between layer 1 and layer 3, layer 1 and layer 5, layer 1 and layer 7 respectively. 
%The network is learning to purify the input information. 

\subsection{Evaluate the Information Inside Channels}
If we further slice $dCNN(x, l, -, t)$ on the channel dimension, we get $dCNN(x, l, c, t)$ which is the collection of all feature maps generated by the same filter $c$. Since filters are the most fundamental building blocks of a CNN model, it is important to investigate the information inside each channel $c$ in a fixed layer $l$ as in 
$dCNN(x, l, c, t)$. When users select one layer in the CNN architecture view and update the channel view, they can visualize the information statistics for each channel of this layer \textbf{(R1, R3)}. To reveal what information get filtered out during the information distillation process, we adopt a Deconvolutional Network (deconvnet) \cite{Zeiler:2011:deconv} to project a filter's output back to the input image space \textbf{(R4)}.

\begin{figure}[htp]
    \centering
    \includegraphics[width=9.5cm]{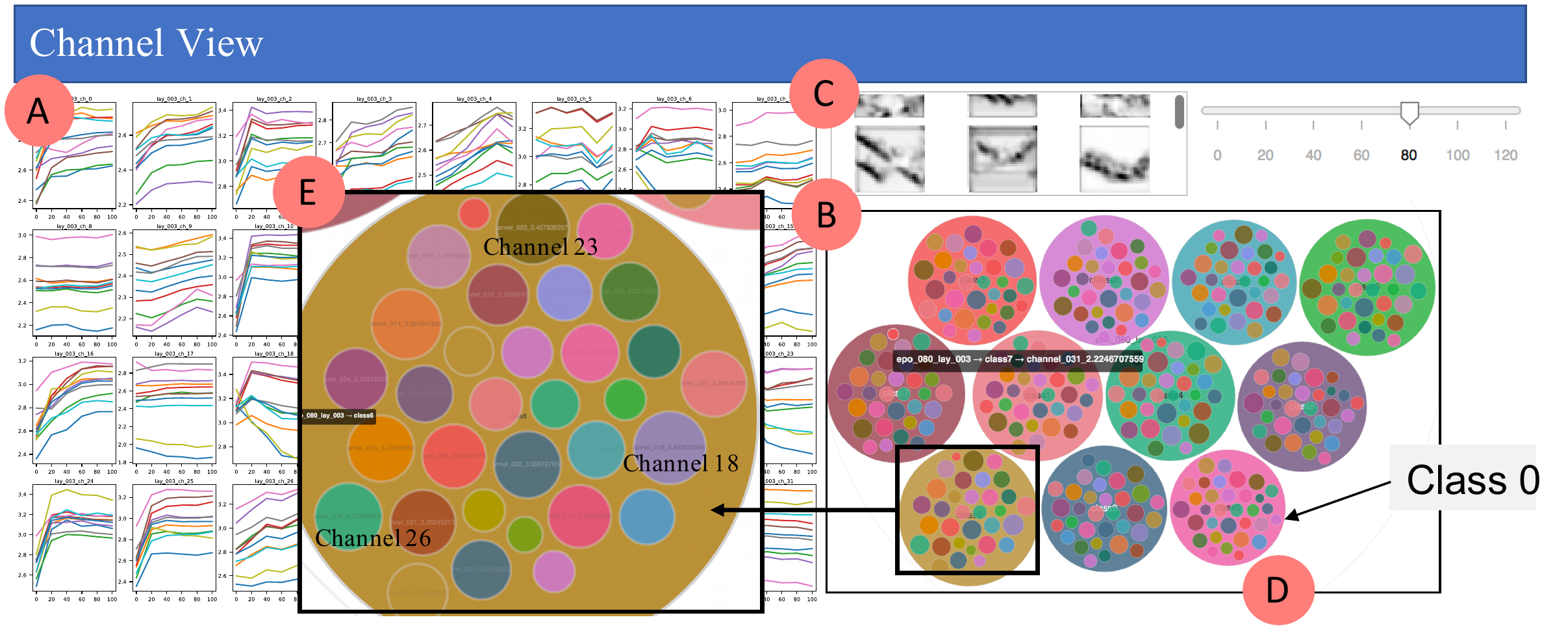}
    \vspace{-12pt}
    \caption{Channel view. (A) Small multiple charts of all channels’ intra-sample entropy over training epochs; (B) circle-packing diagram for all channels in one epoch; (C) deconvolutional results.}
    \label{fig:channelView}
\end{figure}

% For analysis purpose, instead of considering all the data samples we have, it makes more sense if we focus on a subset of them and do comparison, for example, data samples in the same class.
\subsubsection{Information in one channel}
The calculation of $H(dCNN(x, l, c, t))$ is as follows: given a set of input $x$, first we compute the intra-sample entropy for each feature map represented as $H(dCNN(x, l, c, t))$. One feature map corresponds to one input sample, such as as an image. The intra-sample entropy of channel $c$ for all input is the average of all the input's intra-sample entropy. To evaluate a channel's performance for different classes, we compute the channel's intra-sample entropies for all different input classes in a training epoch $t$, meaning we calculate $H(dCNN(x, l, c, t))$ where $x.label=i$ for each $i \in [0, 9]$. 

In the channel view we utilize a small multiple chart as shown in  \autoref{fig:channelView}(A) to visualize all channels' intra-sample entropies' changes across all epochs $t$, where the line graph in each box represents a channel,  the x axis represents epochs, y axis presents the entropy, and different color lines represent different classes. From this small multiple charts, users can quickly locate channels and epochs they are interested in. 

We also employ  circle-packing diagrams with two levels of hierarchy, as shown in \autoref{fig:channelView}(B)to compare intra-sample entropies between different input classes. The input data are divided into subgroups according to their true labels and every subgroup is represented as a circle in the diagram. The color of this first-level circle represents the class. The size of the first-level circle represents the total randomness, i.e., the entropy, of all feature maps belong to this class in this layer. Inside each first-level circle, there are smaller second-level circles representing channels. The size of the second-level circle is proportional to the intra-sample entropy of the channel. Assuming a CNN model has 32 channels in this layer, then there will be 32 second-level circles inside each first-level circle. The color of the second-level circles represents the channel index. Slicing our hypercube dCNN on the epoch dimension can be sliced by a slider. If users are interested in one class in a particular epoch, they can slice the epoch dimension, click on the first-level circle representing this class, the circle-packing diagram can be zoomed in and show the inner second-level circles. By hovering over the second-level circles, users can check the intra-sample entropies for the corresponding channels. 

We utilize a deconvnet which can be thought of as a reversed process of Convolutional Network (convnet) to visualize what information the feature map contains about the input. Deconvnet was used to visualize a trained CNN \cite{Zeiler:2013:deconvCNNvis}. In \cite{Zeiler:2013:deconvCNNvis}, given one input image, feature maps of all channels in a layer are combined and pushed back to the input image space by successively unpool, rectify and filter to get one deconvolutional result. In our work, given one input image, we only use one feature map from the channel we are interested in to get a result in the input image space. Our deconvolutional result can show how much information from the input remains in the selected feature map, i.e., the information distillation result. When users click on one second-level circle, \autoref{fig:channelView}(C) will be updated with the deconvolutional results of the feature maps from this single filter.  

\begin{figure}[htp]
    \centering
    \includegraphics[width=9cm]{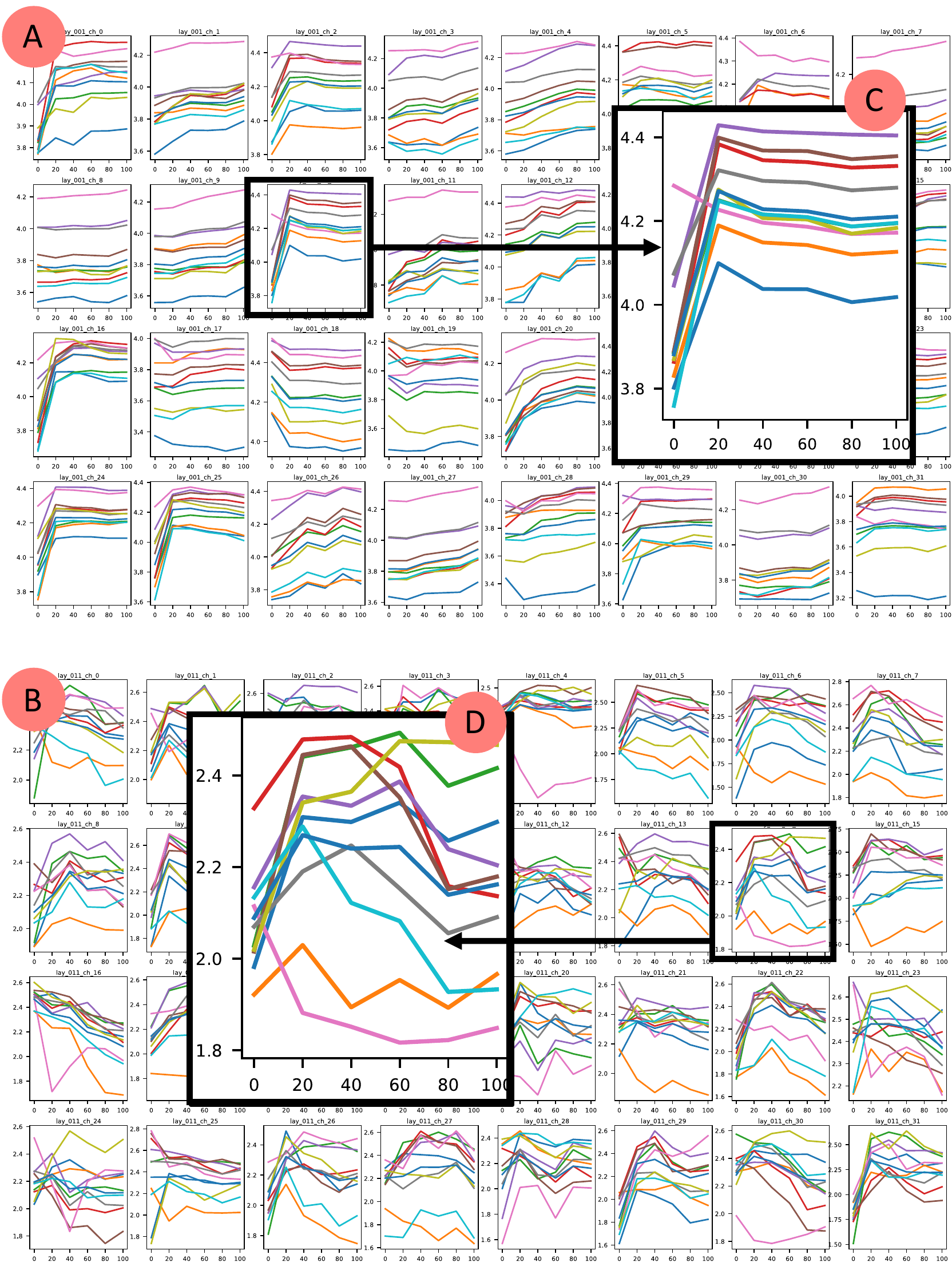}
    \vspace{-12pt}
    \caption{Small multiples charts for layer 1 (A) and for layer 11 (B).}
    \label{fig:ch_l1_l11}
\end{figure}

\subsubsection{Exploration of information inside channels}
We use the CNN classification model in \autoref{fig:queryView}(A) trained on CIFAR-10 dataset for further information investigation. \autoref{fig:ch_l1_l11}(A) is the small multiples chart for layer 1 and \autoref{fig:ch_l1_l11}(B) is for layer 11. By comparing these two plots, we conclude that filters in lower layers are more consistent among all input classes. However, filters of higher layers perform more differently and the behavior depends on the class of the input. This may be because normally filters in lower convolutional layers are trained to extract diverse and detailed features which are common for all classes. But in higher layers, filters need to combine low-level features into class related high-level representations. Besides, since low level feature maps have more high frequency details (contains more irrelevant information), channels in lower layers have relative higher entropies comparing to higher layers as shown in  \autoref{fig:ch_l1_l11}(C) and \autoref{fig:ch_l1_l11}(D).

We are interested in channel's performance differences when classifying different classes. \autoref{fig:channelView}(B) is the circle-packing diagram of layer 3 at epoch 80. Among these 11 big circles, 10 of them representing 10 different classes and the remaining circle representing all feature maps regardless of the class. By comparing the size of each circle, we can get an idea of the the intra-sample entropies when we have different input classes. As shown in \autoref{fig:channelView}(D), class 0, which is the class ``airplane'', has the smallest intra-sample entropies. We also zoom in the circle of class 6 as shown in \autoref{fig:channelView}(E), by comparing the size of circles and hovering over the circles to see the intra-sample entropy values, we locate some channels with higher entropies such as channel 23, 18 and 26. 
\begin{figure}[htp]
    \centering
    \includegraphics[width=7cm]{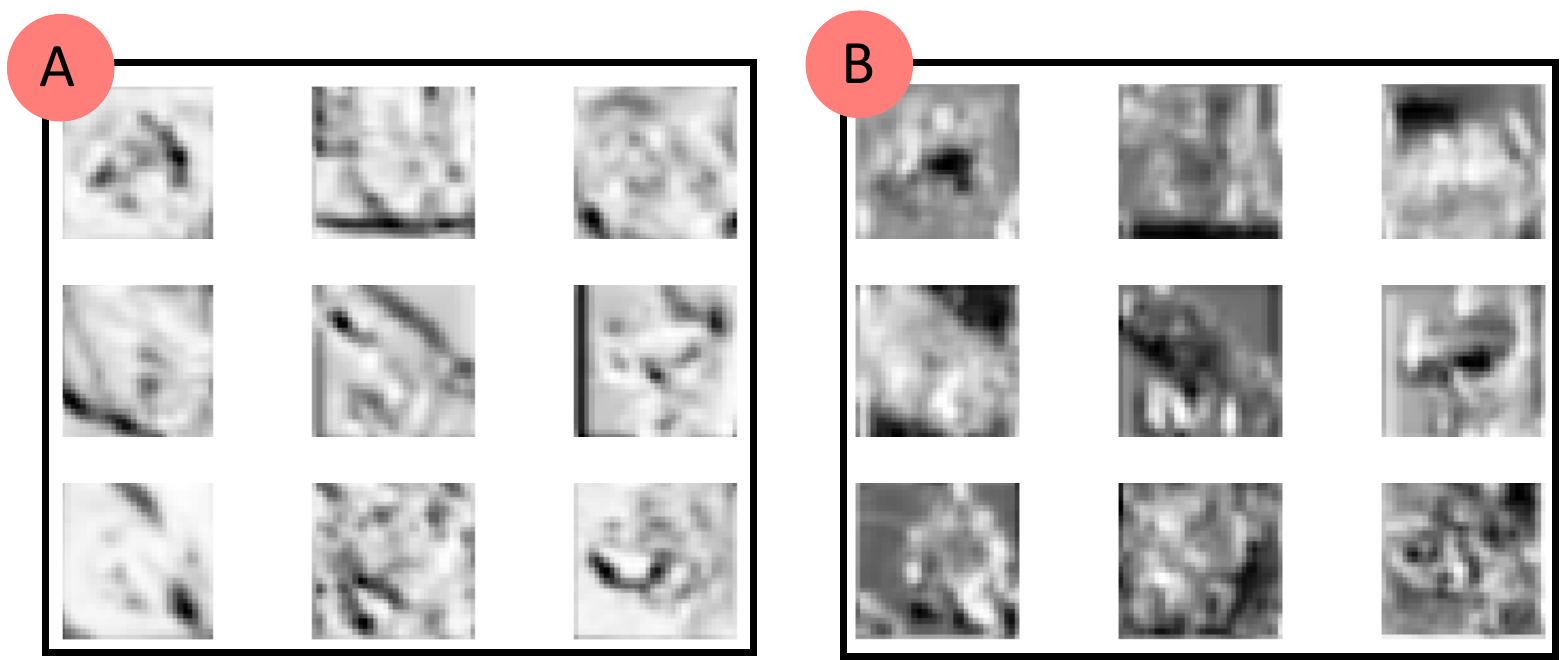}
    \vspace{-6pt}
    \caption{Deconvolutional result for feature maps in channel 31 (A) and channel 23 (B) in layer 3.}
    \label{fig:deconv_f23_31}
\end{figure}
We notice in \autoref{fig:channelView}(E), there is a small second-level circle in the top area near channel 23. By hovering over this circle, we find it is the channel 31 with entropy 1.795243. We find in each class, the size of the circle in this color are all relatively small. To gain insight into the function of the corresponding filter, we adopt deconvnet to visualize this filter's learned features. In class 6, we click on the small circle of channel 31. \autoref{fig:deconv_f23_31}(A) is the deconvolutional result. We also visualize other filters such as filter 23 who has higher inter-sample entropy in \autoref{fig:deconv_f23_31}(B). From \autoref{fig:deconv_f23_31} we can see the result of filter 31 contains less details compared to filter 23. This means filter 31 distills the input information too much. From our case studies, we demonstrate the hierarchy of circle-packing diagram gives users insight into the performance of filters for different classes.

\section{Conclusion and Future Work}
In this work, we combine visualization with an information-theoretic approach to help evaluate and understand the information distillation process of CNNs. To begin with, we formalize a data model to store the information available in a CNN model. A four-dimensional hypercube derived from the  data model, denoted as dCNN(X, L, C, T), consists of four dimensions (i.e., input(X), layer(L), channel(C) and epoch (T)). By slicing on the hypercube, we are able to systematically formulate various of information queries related to CNNs. Based on the analysis need, we propose two types of entropies: inter-sample entropy and intra-sample entropy and show how they can be computed in their respective space. 
We also develop a visual analysis system, CNNSlicer, from which users can explore the flow  of information acorss the various components of a CNN. We use the MNIST and CIFAR-10 datasets to demonstrate how to use our framework to evaluate and analyze the CNN classification model through CNNSlicer, such as comparing the convergence speed of training, and the information changes between layers inside the model.

In the future, we plan to extend our framework to analyzing other types of deep neural networks such as Recurrent Neural Networks (RNN), Generative Adversarial Networks (GAN), and Attention-based transformers. We will also look into  the roles of various neural network components using information theory,  such as the activation functions, batch normalization,  skip connection, and  back-propagation. We will study and compare the landscapes of loss functions for different models using information theory. We believe our information theoretical framework will provide a unique perspective to opening the blackbox of deep learning neural networks. 

% The authors wish to thank A, B, and C. This work was supported in part by a grant from XYZ.}

%\bibliographystyle{abbrv}
% \bibliographystyle{abbrv-doi}
\bibliographystyle{unsrt}

\bibliography{template}
\end{document}